\documentclass[journal]{IEEEtran}

\usepackage{changepage} 
\usepackage{tcolorbox}  
\usepackage{subcaption}  
\usepackage{amsmath} 
\usepackage{float}
\usepackage{graphicx}%
\usepackage{multirow}%
\usepackage{amsmath,amssymb,amsfonts}%
\usepackage{amsthm}%
\usepackage{mathrsfs}%
\usepackage{xcolor}%
\usepackage{textcomp}%
\usepackage{manyfoot}%
\usepackage{booktabs}%
\usepackage{algorithm}%
\usepackage{algorithmicx}%
\usepackage{algpseudocode}%
\usepackage{listings}%
\usepackage{upgreek}
\usepackage[numbers]{natbib}
\usepackage{upgreek}
\usepackage[labelformat=simple]{subcaption}

\newcommand\descitem[1]{\item{\bfseries #1}\\}
\usepackage{enumerate}
\usepackage{graphicx}
\usepackage[justification=centering]{caption}
\DeclareMathOperator*{\argmax}{argmax}

\begin{document}
\title{Towards Futuristic Autonomous Experimentation---A Surprise-Reacting Sequential Experiment Policy}

\author{Imtiaz Ahmed$^{1}$,  Satish Bukkapatnam$^{2}$, Bhaskar Botcha$^{2}$ and Yu Ding$^{3}$
\thanks{This work was supported by NSF grants IIS-1849085 and CNS-2328395.}
\thanks{$^{1}$Imtiaz Ahmed with the Department of Industrial \& Management Systems Engineering,
        West Virginia University,
        Morgantown, WV.
        {Email: \tt\small imtiaz.ahmed@mail.wvu.edu}}%
\thanks{$^{2}$Satish Bukkapatnam with the Dept. of Industrial \& Systems Engineering,
	Texas A\&M,
	College Station, TX.
	{Email: \tt\small satish@tamu.edu}}%
\thanks{$^{2}$Bhaskar Botcha with the Dept. of Industrial \& Systems Engineering,
	Texas A\&M,
	College Station, TX.
	{Email: \tt\small bhaskarb94@tamu.edu}}%
\thanks{$^{3}$Yu Ding with the Stewart School of Industrial \& Systems Engineering,
	Georgia Tech,
	Atlanta, GA.
	{Email: \tt\small yu.ding@isye.gatech.edu}}%
}

\maketitle

\begin{abstract}
An autonomous experimentation platform in manufacturing is supposedly capable of conducting a sequential search for finding suitable manufacturing conditions by itself or even for discovering new materials with minimal human intervention. The core of the intelligent control of such platforms is a policy to decide where to conduct the next experiment based on what has been done thus far. Such policy inevitably trades off between exploitation and exploration. Currently, the prevailing approach is to use various acquisition functions in the Bayesian optimization framework.  We discuss whether it is beneficial to trade off exploitation versus exploration by measuring the element and degree of surprise associated with the immediate past observation.  We devise a surprise-reacting policy using two existing surprise metrics, known as the Shannon surprise and Bayesian surprise. Our analysis shows that the surprise-reacting policy appears to be better suited for quickly characterizing the overall landscape of a response surface under resource constraints. We do not claim that we have a fully autonomous experimentation system but believe that the surprise-reacting capability benefits the automation of sequential decisions in autonomous experimentation.\\

\textit{Note to Practitioners}
Autonomous systems should be able to go beyond repetitive automatic actions that are generally pre-programmed through a \emph{recipe}. To decide what to do next on the fly differentiates autonomy from automation.  Arguably, autonomy is the highest form of automation.  To endow a manufacturing with autonomy, one necessary capability is for it to react properly to the ``unexpected," which are those observations disagreeing with its model's anticipation.  Are these bad measurements, an anomaly, or an indicator of model inadequacy?  Should the observations be discarded or should the model be updated using the new observation?  If latter, should model be updated gradually overtime or radically altered?  Figuratively, upon observing the unexpected, we say that a manufacturing control system is ``surprised" and ask the question of how it should react. Our investigation shares our current insights on this question.  
\end{abstract}


\vspace{-18 pt}
\section{Introduction}

\IEEEPARstart{I}n recent years we witness concerted efforts spent on, and rapid growth in the area of, researching and developing autonomous systems or platforms~\cite{nikolaev:2016,  talapatra:2018, burger:2020, cao:2020, deneault:2021}. An autonomous system is  certainly an automation system.  But researchers use the phrase ``autonoumous'' or ``autonomy'' to emphasize a higher level of automation beyond the repetitive type.  An autonomous system is ideally capable of search a complex design (or parameter) space of process conditions and material elements with minimal human intervention.  Although the above-cited systems are labeled \emph{autonomous}, none in reality is truly or fully autonomous yet.  We think it is more appropriate to label the autonomous systems as ``futuristic".

To help readers understand how an autonomous manufacturing platform should ideally operate, consider an example of laser additive manufacturing (LAM). The platform must navigate a vast combination space of input parameters, such as laser power, scan speed, and layer thickness, to achieve desired properties, like the tensile strength or surface finish, of a manufactured item. An experiment location refers to a specific combination of parameters chosen for a particular action. For example, one experiment location might be defined by the laser power of 200W, scan speed at 300 mm/s, and a layer thickness of 50 microns. Conducting an experiment at this location entails the actions of setting the LAM machine to these parameters, doing the additive manufacturing, and then measuring the resulting properties of the manufactured part.

The goal is to accurately approximate the underlying response surface, which helps understand how different parameter settings impact the manufacturing/materials properties, especially under a limited budget, to quickly reach an ideal setting. Given the large number of parameter combinations, an exhaustive search is impractical due to high costs and time constraints. Therefore, autonomous platforms must employ sequential experimental designs and decision-making processes to identify high-quality settings efficiently and effectively~\cite{lovell:2011}. The key is for the platform's intelligent control to decide where to take action in the next steps and then instruct its hardware system to carry out these actions. In doing so, it always tries to trade off and balance between two actions: exploration and exploitation. Exploitation refers to follow-up investigations in a nearby region by fine-tuning parameters around the current settings, whereas exploration involves conducting experiments in different regions by significantly changing the parameter values to discover potentially better configurations.

Balancing these two approaches is vital, as over-exploitation can miss out on better settings, while over-exploration can waste resources. An ideal method would dynamically adjust this balance based on previous experiments, exploring new parameter spaces when necessary and exploiting known good areas when warranted. This adaptive approach ensures accurate characterization of the response surface within a limited experimental budget, ultimately enhancing the ability to achieve desired material properties in the manufacturing process. While delaying the detailed review of the related literature in Section~\ref{sec:literature}, we would like to point out that the dominating paradigm used nowadays is still the Bayesian optimization (BO) framework~\cite{frazier:2018}, despite the warnings~\cite{bull:2011, chen:2019} that the BO-driven approaches are generally greedy, attempting to hone in on optima too fast and overweighing exploitation over exploration. 

Along the line of finding a better balance between exploration and exploitation, we study the issue of handling the ``surprise observations'' and our research shows that reacting to surprises leads the system to react differently.  Speaking colloquially, a surprise is an observation disagreeing with one's current working hypothesis.  But does a surprise observation mean the hypothesis is wrong or does it mean the observation is corrupted? Pertinent to surprise observations, the following questions need to be addressed: (1) how to define and quantify a surprise? This is to say, given an observation, when should a sequential approach treats it as a surprise and when not? (2) how to react when a surprise is observed? (3) what impact, positive or negative, may there be when a surprise-reacting policy is used? Integrating a surprise measure and devising a corresponding reaction mechanism are considered crucial for advancing autonomous systems~\cite{plataniotis:2022}. 

It turns out that there has been research in the field of information and computer science to provide a quantitative definition of surprise~\cite{baldi:2002, itti:2006, faraji:2018}.  Two widely used ones are the Shannon surprise~\cite{baldi:2002} and Bayesian surprise~\cite{itti:2006}. In this work, we do not intend to introduce new surprise definitions but plan to make use of the existing ones for our analysis. This does not mean that the current surprise definitions are perfect and have no need for improvement (they do need improvement). Rather it just means that as one of the first works on this topic, we choose to dedicate more effort in addressing the other two questions posed above.

Making use of existing surprise measures, we propose a surprise-reacting sequential experimentation policy for materials/manufacturing experiments under a resource-constrained environment. The purpose of these experiments is to get a quick idea about the underlying design space using as few experiments as possible. It falls into the area of ``approximating the underlying function" and different from the objective of finding a single optimal response (a maximum or a minimum). It is not entirely detached from the pure optimization objective, as a good function approximation lays solid ground for optimizing a design in the next step. But understanding the design space provides other benefits. For instance, it can help material scientists or manufacturing engineers to have a holistic view of the entire landscape and thereby enhance the chance to make a new discovery or a robust decision. We contrast our proposed surprise-reacting approach with different acquisition functions~\cite{shahriari2015taking} within the BO framework, which include both exploitation-friendly approaches such as EI and Probability of Improvement (PI) and exploration-oriented acquisition functions such as Upper Confidence Bound (UCB) and Maximum Variance (MaxVar). The performance comparison with these BO-based methods demonstrates the advantage of the proposed surprise-reacting approach.   

What insights do we garner through this research? Those can be summarized in the following two principal aspects.

First of all, taking advantage of surprise observations helps redirect the effort that better balances exploitation and exploration.  Upon observing a surprise, the system spends some extra resources to confirm if the observation is corrupted---an action of exploitation.  If yes, then the current working hypothesis is maintained, and further exploration ensues. If not, the working hypothesis is seriously challenged and the statistical model incorporating the working hypothesis is then updated, so that subsequent exploration will be guided differently. At first glance, the extra resources spent on deciding the nature of an observation is wasteful.  Through our empirical analysis, however, it appears that the additional exploitation informs better subsequent decisions, leading to a better performance than the existing acquisition functions.

Secondly, the introduction of the surprise metric helps explore the underlying design space.  On a high level, we think that the surprised-reacting policy is amounted to an \emph{adaptive} BO, where the adjudication of surprise observations injects adaptivity into the system. It is this adaptivity that helps the learning algorithm escape from local optima and continue searching for new discoveries.

We note that an earlier version of our surprise-reacting idea was posted on \emph{arXiv} and then used in a 3D printing study~\cite{jin:2022}.  In the present study, we refine the idea and present both the benchmark studies and the real-life case differently from what was studied in~\cite{jin:2022}. The rest of the paper unfolds as follows. Section~\ref{sec:literature} summarizes the past relevant research. Section~\ref{sec:algorithm} elaborates on the concept and measurement of surprise and discusses how to react to surprise. We conclude this section by providing a simple illustration example. Section~\ref{sec:performance} presents the performance evaluation of the proposed surprise-reacting approach as compared to the competing approaches.  Finally, we summarize the paper in Section~\ref{sec:conclusion}.

\section{Relevance to the Literature}\label{sec:literature}

Systematic study of experimental designs, or design of experiments (DOE), was initially started with applications to biology and agriculture areas~\cite{fisher:1935}.  Later the DOE methodologies are popularized to many different applications and industries~\cite{wu:2009}. Researchers have long realized the importance of sequential experiments, as it is impossible to understand a complex system fully through a single shot of action.  The early effort of sequential experiments can be traced back to Wald's sequential analysis~\cite{wald:1947}, Box and Wilson's response surface methodology~\cite{box:1951}, and Feldbaum's dual control theory~\cite{feldbaum:1960}. 

The introduction of Gaussian processes (GP) from geo-spatial statistics~\cite{cressie:1991} into the modeling of computer experiment's outputs~\cite{santner:2003} brought a paradigm shift.  GP models were initially used on modeling responses from the \emph{deterministic} computer experiments, which, when run repeatedly with the same input, would produce the same output. GP, being a perfect interpolator, was a natural choice to be used for modeling such responses. Over the years, however, the use of GP models is not limited to the deterministic computer experiments, but also extended to the modeling of stochastic computer simulations~\cite{kleijnen:2008} as well as to modeling physical experiments~\cite{noack:2020, chen:2021, chen:2022}. They become ever more popular when the machine learning era arrives~\cite{rasmussen:2006}.


Jones et al. \cite{jones:1998} proposed the idea of Expected Improvement (EI), a criterion to decide where to collect the next data point. The EI criterion tries to balance between sampling the next data point with the highest expected value (exploitation) and sampling the point with the highest uncertainty (exploration) and was proven effective. The effort of finding a GP-driven efficient global optimization is evolved into the research of Bayesian optimization. BO decides the next sample point by optimizing an acquisition function and EI forms one such function.  Other popular choices of acquisition function include Probability of Improvement~\cite{kushner:1964}, Upper Confidence Bounds~\cite{cox:1992, srinivas:2010}, and Maximum Variance~\cite{mockus:1978}. Recent years one has seen new applications of BO in areas such as manufacturing~\cite{gongora:2020, albahar:2021}, material handling~\cite{kang:2023}, neuroscience~\cite{lancaster:2018}, and materials discovery~\cite{zhang:2020}.



Recent research~\cite{wang2023recent, ghorbani2024active,di2024active} has highlighted the limitations of various BO acquisition functions, particularly their tendency to either over-exploit or over-explore the search space. For instance, EI is known for its exploitation-heavy nature, often leading to local optima without sufficient exploration of the broader design space. UCB aims to balance exploration and exploitation, but its effectiveness heavily depends on the choice of hyperparameters, which may not be adaptive enough for dynamic environments. PI and MaxVar face similar challenges, where PI tends to focus on regions with high expected improvement and MaxVar targets areas of high uncertainty without necessarily finding optimal solutions. Adaptive strategies are identified as the best way to balance exploration and exploitation to improve performance in varying conditions~\cite{han2023adaptive}. The dynamic adjustment of acquisition functions in real-time is demonstrated to better suit the needs of autonomous platforms~\cite{islam2024dynamic}. These findings underscore why our surprise-reacting policy could be advantageous, for it adjusts exploration and exploitation dynamically based on the degree of surprise and is thus adaptive, offering a more robust and effective approach for autonomous experimentation.


We have previously mentioned a number of existing autonomous platforms, although none is fully autonomous yet. Most of these platforms still rely on BO to direct their choices in the sequential experimentation, including the \emph{robotic chemist}~\cite{burger:2020}, featured in a \emph{Nature} cover story. Bukkapatnam~\cite{bukkapatnam:2022} provides a comprehensive overview on autonomous manufacturing, stressing that a lot of challenges remain unsettled and that we are still far away from developing a practical sequential strategy for these autonomous platforms. It is in this context that we would like to report our work in terms of incorporating the element of surprise into a sequential experimentation policy.

\section{Surprise-Reacting Experimentation Policy}\label{sec:algorithm}


This section discusses how to handle surprise observations, i.e., how to measure a surprise and how to react to it. The reaction plays the role of guiding the sequential framework to select the next experiment location and update its understanding of the design space or the underlying response function. 

Before we proceed with the technical discussion, we would like to present a brief account of how ``surprises'' acted as an important element in the process of scientific discovery. Surprise can be considered as the observations that disagree with the current hypotheses concerning or the understanding of the underlying systems. Surprise often brings forth puzzlement first, and as an immediate reaction, one investigates further to understand the surprise. It may trigger an adjustment to one's current understanding of the systems (or processes) and eventually leads to the sublime knowledge one aspires to reach, known as enlightenment. Whenever one is surprised, a natural scientific feedback is to exploit the neighborhood, close to the surprise location, to find out the nature and extent of the surprising responses. This is the process of adjudicating the surprise observation, and it could help unearth a new feature or pattern of the response surface.

Surprise can be linked to two different states of mind, i.e., puzzlement and enlightenment, that comes one after another~\cite{faraji:2018}. To appreciate the two different states, let us revisit the discovery process of penicillin by sir Alexander Fleming in 1928, a bacteriologist working at St. Mary's Hospital in London. One day upon returning from a two-week summer vacation, sir Alexander found that a culture plate of \emph{Staphylococcus aureus} that he had been working on was contaminated by a mold, which inhibited the growth of the \emph{Staph} bacteria. He was puzzled by this outcome and named the mold broth filtrate penicillin. He did not stop there or threw away the culture plate as a ``bad'' data point but started investigating the event instead, which led to the discovery of antibiotic. The two states of mind of Sir Alexander were: the puzzlement in the first hour right after being surprised (when bacteria growth was stopped) and the enlightenment when he understood the reason (discovery of antibiotics).

\subsection{Measures of Surprise}

To learn from surprise, a well-defined mathematical measure is needed to quantify the abstract concept. As we explained in the introduction, we intend to use the two existing surprise definitions known as Shannon surprise~\cite{baldi:2002} and Bayesian surprise~\cite{itti:2006}, respectively.

\vspace{5 pt}

\textbf {Shannon surprise} uses a negative log-likelihood of an observation, $\mathbf D=\{\mathbf x, y\}$, given the current state of mind. Here, $\mathbf {x}$ represents the input event while $y$ represents the observed response. Let us use $\pi_{n}(\boldsymbol \uptheta)$ to represent the current state of mind, i.e., the belief regarding the underlying system captured by a statistical model (more on this in the next subsection) after observing $n$ data points. This state of mind is parameterized through $\boldsymbol \uptheta$. Then the Shannon surprise is defined as:
\begin{equation}
\label{eq:Sh}
-\log \int_{ \boldsymbol\uptheta}p( \mathbf D \mid \boldsymbol\uptheta)\pi_{n}(\boldsymbol \uptheta)d\boldsymbol \uptheta,
\end{equation}
where $p(\mathbf D \mid\boldsymbol \uptheta)$ measures the probability of a new data point, conditioned on $\boldsymbol \uptheta$. The degree of surprise is proportional to the value of the Shannon surprise measure. Observations with a low probability of occurrence imply a big surprise.

Shannon surprise measures a surprise using the posterior probability. However, not all low-probability events are surprising. Faraji et al. \cite{faraji:2018} uses the following example to illustrate the point.  Consider that someone noticed a car of a specific make, a specific color, and a specific license plate parking next to his/her own car in a parking lot. Assume that all cars are parked randomly. Given so many cars out there, the probability of observing a particular car parking next to one's own is very low.  Yet, one will not be typically surprised by this low probability event because one does not have the anticipation of either seeing or not seeing that car in the first place. In Shannon surprise, neither does one update his/her belief after seeing the surprising event, nor does one compare it with the prior belief. In other words, Shannon surprise does not capture the sense of anticipation.

\vspace{5 pt}

\textbf {Bayesian surprise}: It captures the change in one's belief brought by the newly observed data point. Bayesian surprise quantifies the change using the Kullback-Leibler (KL) divergence between the distribution of the prior belief and that of the posterior belief, such as:
\begin{equation}
\label{eq:BS}
KL(\pi_{n}(\boldsymbol \uptheta)\parallel \pi_{n+1}(\boldsymbol\uptheta)), \quad \text{and}
\end{equation}

\begin{equation}
\label{eq:BS1}
\pi_{n+1}(\mathbf \uptheta)=
\frac{p(\mathbf D\mid \boldsymbol\uptheta)\pi_{n}(\boldsymbol\uptheta)}{\int_{ \boldsymbol\uptheta}p(\mathbf D\mid \boldsymbol \uptheta)\pi_{n}(\boldsymbol\uptheta)d\boldsymbol \uptheta}.
\end{equation}

Here, $\pi_{n+1}(\boldsymbol \uptheta)$ represents the updated belief after observing a new data point ($\mathbf D$) and is calculated using the Bayes rule.  Bayesian surprise updates the state of mind after observing new data; that is an act of enlightenment. The KL divergence compares the two believes, and the prior belief serves as the anticipation. Events that cause a big change in one's belief, i.e., a big KL divergence, are labeled as surprises. Apparently, Bayesian surprise is a measure of the enlightenment surprise.

Compared to the Bayesian surprise, Shannon surprise is more sensitive and can lead to a faster reaction, because Shannon surprise is about capturing the initial puzzlement, whereas Bayesian surprise is more about updating the belief. Consider the example that a student with a very good grade history suddenly gets a poor grade in one test. Using the Shannon measure, under the belief, $\pi_n(\boldsymbol \uptheta) = \{\text{good student}\}$, the probability of $\mathbf D =\{\text{bad score}\}$ is low, so that one would be puzzled and tag this event as a surprise. On the contrary, using the Bayesian measure, one is unlikely to change from $\pi_n(\boldsymbol \uptheta) = \{\text{good student}\}$ to $\pi_{n+1}(\boldsymbol \uptheta) = \{\text{bad student}\}$ after one single test. This is to say,  $\pi_{n}(\boldsymbol \uptheta)$ and  $\pi_{n+1}(\boldsymbol \uptheta)$ stay the same and the KL divergence is close to zero. As such, one is not surprised if using the Bayesian measure.  In order to change the belief, much more bad scores are needed to gradually overturn the prior belief of ``\{good student\}''.  The slower response often comes as a criticism of Bayesian surprise, especially when the experiments are expensive and resources are precious (meaning that one does not afford a lot of new observations to react).

It appears that neither of the existing surprise measures are perfect~\cite{faraji:2018}.  While introducing a new surprise measure is worthy, doing so is not straightforward.  We believe the existing measures, however imperfect they may be, are still useful.  So we stay with the two definitions and demonstrate their usefulness in this paper.

\vspace{-6 pt}

\subsection{Statistical Model for Sequential Experimentation}\label{sec:model}

In order to compute the above-defined surprise measures and incorporate them into a sequential framework, we need to introduce a statistical model. We expect this model to hypothesize one's belief over the design space and sequentially update its belief by using the new observations. In this work, we choose to adopt the Gaussian process as the statistical model. Using GP makes our effort better connected with and comparable to the work under the BO framework. 

\subsubsection{GP Model}
Consider we are trying to model physical experiments with an underlying function \(f(\mathbf{X})\). Let \(\mathbf{X} = [\mathbf{x}_1, \mathbf{x}_2, \ldots, \mathbf{x}_n]\) represent the observed experiment locations. GP models the function values at these locations as a vector of random variables, \(\mathbf{f} = [f(\mathbf{x}_1), f(\mathbf{x}_2), \ldots, f(\mathbf{x}_n)]\) and assumes that the set of random variables in \(\mathbf{f}\) is jointly distributed as a multivariate Gaussian:

\begin{equation}
\label{eq:GP}
GP \sim p(\mathbf{f} \mid \mathbf{X}) = \mathcal{N}( \boldsymbol{\upmu}, \mathbf{K}),
\end{equation}
where \(\boldsymbol{\upmu} = m(\mathbf{X}) = [m(\mathbf{x}_1), m(\mathbf{x}_2), \ldots, m(\mathbf{x}_n)]\) is the mean function, and \(\mathbf{K} = [k(\mathbf{x}_i, \mathbf{x}_j)]_{i,j=1}^n\) is the covariance or kernel function.

The mean function defines the expected value of the process at any point. It is often set to zero (i.e., \(m(\mathbf{X}) = \mathbf{0}\)) for simplicity, particularly when there is no prior information about the trend in the data. Even with this simplistic choice, GPs, with their flexible covariance functions, can model complex patterns and deviations from the mean well. GPs can also incorporate prior knowledge by using non-zero mean functions, such as linear, polynomial, or other parametric forms, to reflect known trends or biases in the data. 

The covariance function, or kernel, \(k(\mathbf{x}_i, \mathbf{x}_j)\), dictates the structure and smoothness of the function being modeled in a GP. Generally, the choice of kernel determines how the function values correlate with each other. There are two broad categories of kernels: stationary and non-stationary. Stationary kernels, such as the squared exponential, Matérn, and Ornstein-Uhlenbeck functions, depend only on the distance between \(\mathbf{x}_i\) and \(\mathbf{x}_j\) and not on their absolute locations. This implies that the covariance between function values is a function of the relative distance. Non-stationary kernels, on the other hand, can vary with the absolute positions of the points, allowing for more flexibility in capturing varying structures in different regions of the input space.

GP assigns a prior on the function space that the sequential algorithm tries to master using the data from sequential experiments. The covariance function ($\mathbf{K}$) plays the most significant role in this prior as it encodes the similarity of each pair of experimental data points. In this study, we utilize the Matérn kernel due to its flexibility in modeling different degrees of smoothness:
\begin{equation}
\label{eq:se}
k(\mathbf{x}_i, \mathbf{x}_j) = \sigma_s^2 \frac{2^{1-\nu}}{\Gamma(\nu)} \left(\sqrt{2\nu} \frac{\left\Vert \mathbf{x}_i - \mathbf{x}_j \right\Vert}{l}\right)^\nu K_\nu \left(\sqrt{2\nu} \frac{\left\Vert \mathbf{x}_i - \mathbf{x}_j \right\Vert}{l}\right),
\end{equation}
where $K_{\nu}$ is the modified Bessel function, $\nu$ controls the smoothness of the learned function.  We use $\nu=2.5$ throughout our study. There are two parameters in this function whose values need to be learned during the sequential learning process: \(l\) is the length-scale parameter controlling the smoothness of the functional representation of the design space and \(\sigma_s\) is the variance parameter determining the magnitude of the function values. 

The actual experiments may be noisy, which means the observed response, $y$, can contain variability not captured by the underlying function. In GP, noise in the observations can be modeled as either independent and identically distributed (i.i.d.) Gaussian noise or as non-i.i.d. noise. Using i.i.d. noise  simplifies the modeling but may not capture more complex noise structures present in real experiments.

For this study, we choose to model the noise as i.i.d. Gaussian noise for simplicity and computational efficiency. The observed output \(y\) at any input point \(\mathbf{x}\) is given by:
\begin{equation}
y = f(\mathbf{x}) + \xi, \quad \xi \sim \mathcal{N}(0, \sigma^2),
\end{equation}
where \(\sigma^2\) is the variance of the noise, reflecting constant observational uncertainty across the design space. This \(\sigma^2\), along with the parameters of the chosen covariance function (\(l\), \(\sigma_s\)), also known as GP hyperparameters and represented by \(\boldsymbol{\phi}\), constitute the parameters of the statistical model.

The choice of hyperparameters is part of the modeling effort and often utilizes domain or expert knowledge. We do not consider any informative prior distribution on these hyperparameters. Instead, we use flat priors, i.e., start with arbitrarily selected hyperparameter values (or values informed by domain knowledge, bounded within a minimum and maximum range), and use data to iteratively update and optimize these hyperparameters after each iteration.

The GP model is utilized as a surrogate for the true underlying function of the design space, enabling updates through a probabilistic Bayesian learning process. Initially, a set of experiments, \((\mathbf{X}, \mathbf{y})\), where \(\mathbf{y} = [y_1, y_2, \ldots, y_n]\), provides a basic understanding of the design space. The model then updates its parameters by maximizing the log marginal likelihood.
:
\begin{equation}
\begin{aligned}
\label{eq:mar}
\log p(\mathbf{y} | \mathbf{X}) &= -\frac{1}{2}(\mathbf{y}-m(\mathbf{X}))^\top \mathbf{K}_y^{-1} (\mathbf{y}-m(\mathbf{X})) \\
&\quad -\frac{1}{2}\log |\mathbf{K}_y| -\frac{n}{2} \log 2\pi,
\end{aligned}
\end{equation}
where $\mathbf{K}_y = \mathbf{K} + \sigma^2 \mathbf{I}$, with $\mathbf{I}$ being the identity matrix and $\left | \mathbf K_{y} \right |$ represents the determinant of $\mathbf K_{y}$.

To measure surprise, it is important to predict the experimental outcomes at new test locations (\(\mathbf{X}_{*}\)) so that the predicted values can be compared with the actual outcomes (\(\mathbf{y}_{*}\)). This comparison helps to update the sequential policy after each iteration. The prediction is made easier due to the GP formulation, for which the posterior predictive distribution of the function response at the new locations, \(\mathbf{f}_{*}\), is well known to be~\citep{rasmussen:2006}:
\begin{equation}
\label{eq:pr}
p(\mathbf{f}_{*} \mid \mathbf{y}, \mathbf{X}_{*}, \mathbf{X}) \sim \mathcal{N}(\boldsymbol{\mu}_{*}, \boldsymbol{\Sigma}_{*}),
\end{equation}
where
\begin{equation}
\label{eq:var}
\boldsymbol{\mu}_{*} = \mathbf{K}_{*}^{T} \mathbf{K}_{y}^{-1} \mathbf{y} \quad \text{and} \quad \boldsymbol{\Sigma}_{*} = \mathbf{K}_{**} - \mathbf{K}_{*}^{T} \mathbf{K}_{y}^{-1} \mathbf{K}_{*},
\end{equation}
and \(\mathbf{K}_{*} = k(\mathbf{X}, \mathbf{X}_{*})\) and \(\mathbf{K}_{**} = k(\mathbf{X}_{*}, \mathbf{X}_{*})\). This posterior distribution, \(p(\mathbf{f}_{*} \mid \mathbf{y}, \mathbf{X}_{*}, \mathbf{X})\), reflects the current understanding of the design space and can be sequentially updated with each new experiment.

\vspace{\baselineskip}
\subsubsection{Capturing and Labeling Surprise}

We can compute the Shannon surprise of a new observation, $\mathbf D=\{\mathbf x_{*}, y_{*}\}$, by using the GP posterior predictive distribution. Note that the state of mind, $\boldsymbol \uptheta$ is equivalent to $\mathbf{f}$ of the GP model, and the current state of mind, $\pi_{n}(\boldsymbol \theta)$ is equivalent to the GP posterior $p(\mathbf f_{*}\mid y, \mathbf X_{*}, \mathbf X)$, as in Equation~\eqref{eq:pr}, after observing $n$ data points.

To measure the Bayesian surprise, first, we need to include the new experiment location and its response to our GP model's training dataset. The model hyperparameters ($\boldsymbol \phi$) will be then re-estimated through the optimization formulation as in Equation \eqref{eq:mar}, so that we get an updated posterior distribution through Equations \eqref{eq:pr} and \eqref{eq:var}. This updated model ($\pi_{n+1}(\boldsymbol \uptheta$)) will be compared with the old model ($\pi_{n}(\boldsymbol \uptheta)$) through their KL divergence to measure the surprise.

To label a new observation as a surprise, a threshold is needed. We propose to use the credible interval ($\boldsymbol \upmu_{*}\pm k_{\text{Shannon}}\boldsymbol \Sigma_{*}$) for each new test response ($y_{*}$). For instance, when the credible interval is set at 95\%, $k_{\text{Shannon}}=1.96$ for a normal distribution. This credible interval can be easily computed given the posterior distribution. If the degree of surprise associated with an observation is greater than the degree of surprise associated with the credible band, it implies that the new test response does not agree with the statistical model and then this new experiment will be treated as surprise.

Similarly, Bayesian surprise value will be compared with $k_\text{Bayesian}$, which is the counterpart of $k_\text{Shannon}$ above, for deeming a surprise.  When $k_\text{Bayesian}$ is chosen to be the same as $k_\text{Shannon}$, using Bayesian surprise leads to a slower reaction, or alternatively, in order for Bayesian surprise to have a comparable rate of reaction as that using Shannon surprise, $k_\text{Bayesian}$ generally needs to be smaller than $k_\text{Shannon}$. In Section~\ref{sec:performance}, we analyze the performance variations with different threshold selections and discuss the resulting impacts.

\subsection{How to React to Surprise}

Once an observation is flagged as a surprise, the next question is how the model should react to that declaration. To mimic a human scientist, the next action is to investigate the nature of the surprise observation. We understand that the nature of the investigation could vary, but given our focus on engineering systems for manufacturing and material discovery, our experience suggests that the first line of action is to confirm whether the surprise is due to data corruption or a discrepancy between the underlying response surface and the model that has been built thus far.

While settling on such a question itself entails complexity, we choose to conduct a simple test for the time being. The test is for the sequential approach to draw a new observation in close proximity to the location where the surprise is declared and see if it is a surprise again. In this context, we define ``close proximity" as a small perturbation around the last experimented point within the Euclidean space, using a normal distribution with a small standard deviation. Mathematically, for a given point \(\mathbf{x} \in \mathbb{R}^d\), where $d$ is the dimension of the input space, we generate a point \(\mathbf{x}_{\text{perturbed}} \in \mathbb{R}^d\) as follows:
\begin{equation}
\label{eq:expl}
\mathbf{x}_{\text{perturbed}} = \mathbf{x} + \boldsymbol{\epsilon},
\end{equation}
where \(\boldsymbol{\epsilon} \sim \mathcal{N}(0, \sigma_{\text{perturb}}^2 \mathbf{I})\) and \(\sigma_{\text{perturb}}\) is a small standard deviation that defines the ``neighborhood" within the Euclidean space.

If the response at the perturbed location is once again a surprise, it confirms the earlier finding. Both responses are kept in the data collected and used to update the model to reflect the new understanding. If the model is not surprised by the subsequent response, it suggests that the previous surprising observation is more likely a corrupted observation. Then, the previous observation is discarded and not used to update the model.

The confirmation process is an act of exploitation, as it is conducted within a defined local region. This action commits additional resources to double-checking, which on the surface slows down the experiment progress. However, we find that a simple exploitation action like this actually helps with the overall exploration of the design space. The surprises serve as wake-up calls to prevent misleading new data from dragging the model to incorrect conclusions. This is particularly critical in resource-constrained processes, such as running experiments in material sciences or manufacturing, where high costs limit the number of experiments.


After the confirmation step, additional observations would be taken at the same ``neighborhood" for model updating, until a new data collection does not return a surprise declaration.  What this means is that the model is now consistent with what the data informs the model for that local area.  What this entails in reality is just one or two additional data points, not an undue burden for the overall experimentation process.

Once the model is updated to accurately reflect the local area just being exploited, the sequential learning process goes back to the exploration mode, which is to look for new patterns elsewhere in the design space which in principle should be far away from the previously experimented locations. During each exploration step of the sequential experiment, we initialize a set of candidate points, denoted by \(\mathbf{S}\). The new experiment location, \(\mathbf{x}_{\text{next}}\), will be selected from this set based on the set of previously experimented locations, denoted by \(\mathbf{E}\). 

On the note of candidate points, we want to emphasize that our model does not strictly follow BO but rather employs a Bayesian mechanism for sequential experimentation. Unlike traditional BO, which typically optimizes an acquisition function to select the next experimental location, our approach leverages surprise-driven Bayesian inference to guide the selection of the next points for exploration or exploitation. The objective of our method is to quickly approximate the design space and obtain an overall understanding, as opposed to just optimizing expensive black-box functions as in BO. The use of candidate points balances computational efficiency and effective exploration of the design space and thus helps achieve our objective. One more advantage for using the candidate sets is that doing so can incorporate complex design constraints more easily; for example, infeasible designs or designs known to be harmful. 

We generate the candidate set \(\mathbf{S}\) at each iteration using Sobol sequences~\cite{sobol1967} as in Equation~\eqref{eq:expp}, which provide a well-distributed set of points in the Euclidean space. 

\begin{equation}
\label{eq:expp}
\mathbf{S} = \text{Sobol}(\mathbf{X}_{\text{bounds}}, n_{\text{candidates}}),
\end{equation}
where \(\mathbf{X}_{\text{bounds}}\) defines the bounds of the input space and \(n_{\text{candidates}}\) is the number of candidate points.

Sobol sequences are a type of low-discrepancy sequence used in quasi-random number generation and are particularly useful in high-dimensional integration and optimization problems because they provide better space-filling properties compared to purely random sequences. This ensures that the candidate points cover the design space more uniformly, making them ideal for exploration in Bayesian optimization. In this work, we generate 5,000 new candidate points in each iteration. For higher dimensional ($d>6$) and more complex functional spaces, more points may be generated to maintain good coverage.

To select the next experiment location, we use a maximin strategy, which seeks to maximize the minimum distance from the new location to all previously explored locations; see Equation~\eqref{eq:exp}. This strategy ensures good coverage of the design space by preventing clustering of points and promoting exploration~\cite{johnson:1990, morris:1995, joseph:2015}. Specifically, to identify the nearest neighbour of experimented locations in the candidate set, we use Ball Trees~\cite{omohundro1989} for efficient distance computations. Ball Trees are used for efficient nearest-neighbor queries which organize points in a hierarchical structure based on their distances. This allows for fast distance computations and nearest-neighbor queries, which is crucial for our maximin strategy.

\begin{equation}
\label{eq:exp}
\mathbf{x}_{\text{next}} = \arg\max_{\mathbf{x} \in \mathbf{S}} \left( \min_{\mathbf{e} \in \mathbf{E}} \text{BallTree}(\mathbf{x}, \mathbf{e}) \right),
\end{equation}

The degree of surprise associated with the new observation will then be evaluated. In the event of no surprise, the exploration will be continued until a surprise is encountered. Upon encountering a surprise, exploitation will begin. This iteration will be repeated until the experimentation budget is exhausted.


Using this surprise-reacting policy, the expected number of surprises decreases rather quickly as the experimentation proceeds, because the underlying response is better understood. The sequential approach is able to recuperate the benefit of not being stuck in a local neighborhood.  As a result, the surprise-reacting policy could approximate the design space quickly.

\subsection{Steps of the Surprise-Reacting Sequential Experimentation}
\begin{figure}[t]
\begin{center}
\includegraphics[scale=0.5]{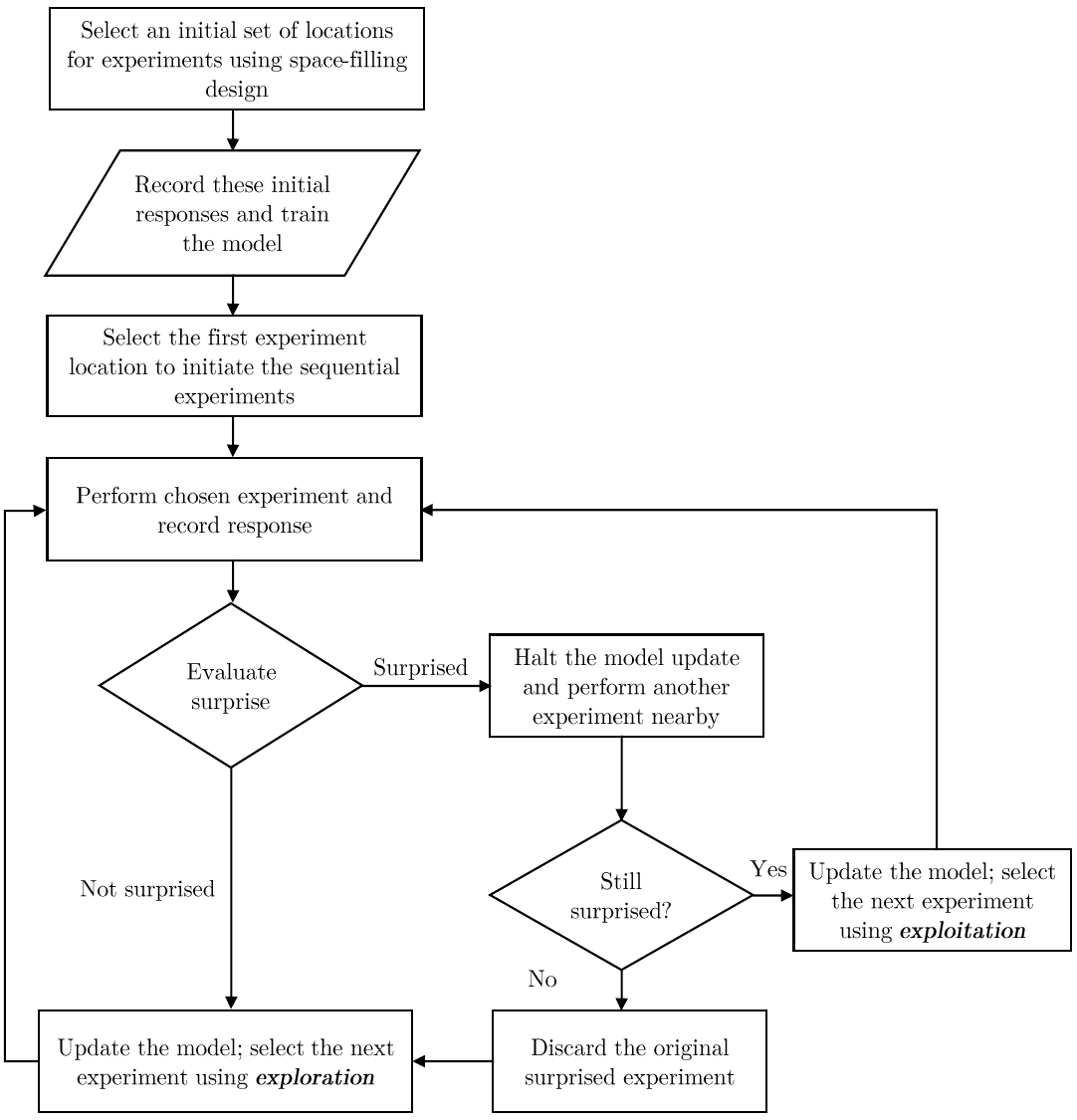}
\caption{The flowchart of the surprise-reacting experimentation policy.} \label{Figure2}
\end{center}
\end{figure}

The policy is described step wise below, which is also summarized in Fig.~\ref{Figure2}.

\begin{enumerate}[I.]

\descitem{\textbf {Initial experiments}:} In order for the sequential experiment to begin, the sequential algorithm must be given a set of initial experimental locations ($\mathbf {X}_{\text{initial}}$) and the corresponding responses, in order to build a statistical model.  For this purpose, we still use the Sobol sequence as we do for the sequential experimentation. The number of experiments in $\mathbf {X}_{\text{initial}}$ are required to be low compared to the total allocated experimentation budget. Considering the resource constrained nature of the autonomous experimentation platform, in this work, we use only 10 initial experiments for all the benchmark functions and the grinding dataset. Once the responses from these experiments, $\mathbf {y}_{\text{initial}}$, are recorded, the statistical model will be trained using these initial experiments, which can then be used to produce the posterior predictive distribution. 

\descitem{\textbf {Surprise measure}:} After the initial experiments, the very next experiment location will be randomly selected, $\mathbf {x}_{\text{first}}$.  The experiment will be carried out to get the response of $y(\mathbf {x}_{\text{first}})$. The model posterior predictive distribution is used to calculate the Shannon surprise (using Equation \eqref{eq:Sh}), whereas the Bayesian surprise is calculated by comparing the old and updated model (using Equation \eqref{eq:BS}). Then, the statistical model is updated by adding this new location, i.e., $\mathbf X=\{\mathbf {X}_{\text{initial}},\mathbf {x}_{\text{first}}\}, \mathbf y=\{\mathbf {y}_{\text{initial}}, y_{\text{first}}\}$. 

\descitem{\textbf {Exploration-Exploitation switching}:} If the sequential algorithm is not surprised, exploration will be pursued. The next exploration location is chosen according to the maximin criterion, i.e., following Equation \eqref{eq:exp}. On the other hand, if the sequential algorithm is surprised, the next location will be selected according to our exploitation policy, i.e., following Equation \eqref{eq:expl}.

\descitem{\textbf {Update}:} Once the next experiment location is selected, actual experiments will be performed. The update to the statistical model depends on the outcome of the exploitation of a surprise observation. The sequential algorithm then moves back to Step II and continues until the allotted experiment budget is reached.

\end{enumerate}

\subsection{A Simple Illustrative Example}

To highlight the differences between the proposed approach and the existing EI-based BO sequential policy, we would like to walk through a simple function approximation problem.

\subsubsection{Problem Description}\label{sec3.5.1}
\begin{figure}[t]
\begin{center}
\includegraphics[scale=0.5]{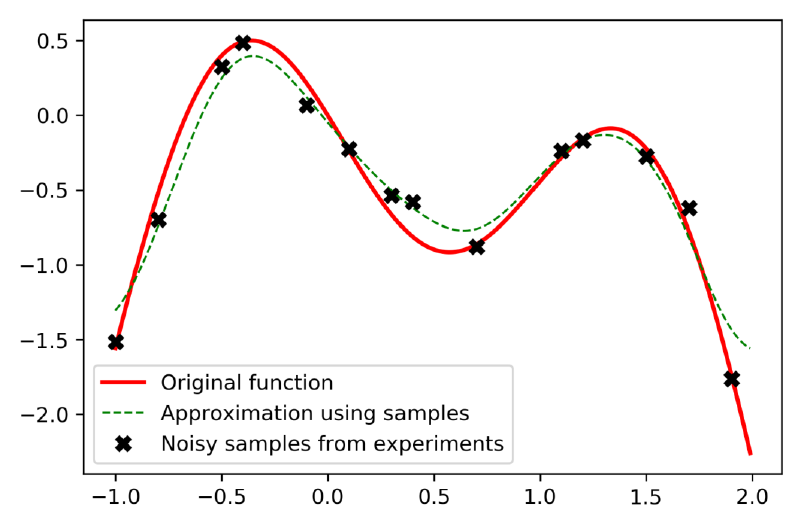}
\caption{Approximation of a response function.} \label{approx}
\end{center}
\end{figure}

In this example, we consider a simple, univariate function of the form of
\begin{equation}
\label{eq:response}
y = f(x) + \upxi.
\end{equation}
The specific $f(\cdot)$ used in this simple example is $-\sin (3x) -x^{2}+0.7x$, shown as the solid red curve in Fig. \ref{approx}. One can observe that the $f(\cdot)$ function has two peaks, one higher than the other.  Without knowing the underlying true function, a sequential algorithm would run experiments and take responses. Those are marked as the black crosses in the figure. Once there are a sufficient number of data pairs, the statistical model can recover the function reasonably well, which is the dotted green curve. For a simple function of a single input as in Fig. \ref{approx}, one does not need a large number of experiments before recovering the underlying true function. This example is simply used for illustration purposes.

Specifically in this example, the realization of $y$ is the addition of function $f(\cdot)$ with a zero-mean Gaussian noise with $\sigma=0.2$.  The experimental budget is constrained to 13 physical experiments, including the two initial experiments. Input $x$ takes value in the range of $[-1.0, 2.0]$. We use both the lower and upper bound values as the initial experiment locations; these are the same for both the surprise-reacting policy and the EI-based BO policy. The sequential experiments start after these two initial experiments and run a total of 11 additional experiments which makes in total 13 experiments. We use a GP with Mat\'ern kernel as the statistical model for all policy options. We use a pre-determined lengthscale parameter value of 1 and a smoothness parameter value of 2.5.

\subsubsection{Surprise-reacting experiment policy }
\begin{figure*} [!htb]
\begin{adjustwidth}{-2cm}{-2cm} 
\centering
        \begin{subfigure}{0.43\linewidth}
        \centering
        \caption{Approximation using Shannon surprise.}
        \label{fig:M1}
        \includegraphics[width=\linewidth]{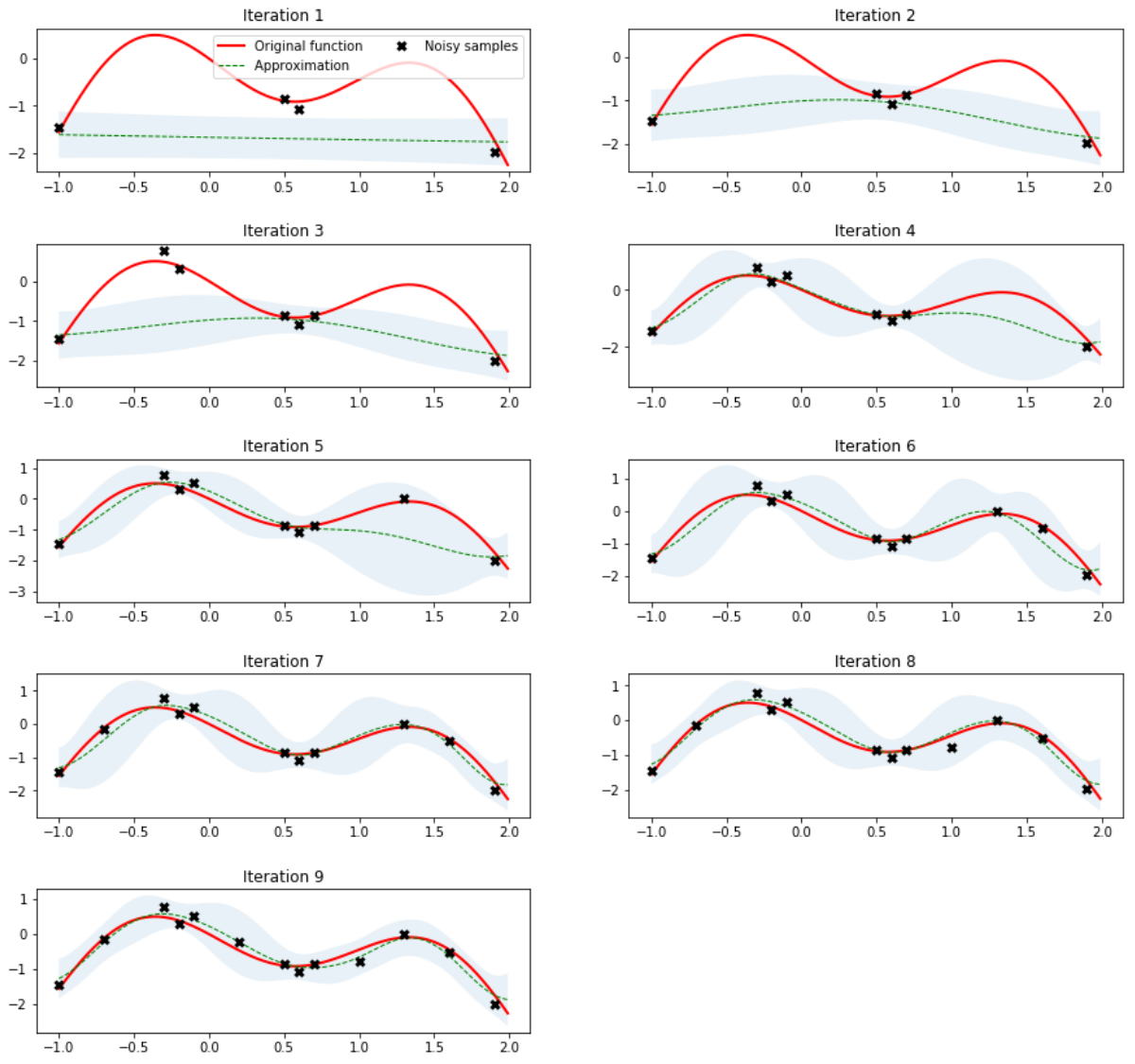}
    \end{subfigure}%
    \hspace{5 pt}
    \begin{subfigure}{0.43\linewidth}
        \centering
        \caption{Approximation using Bayesian surprise.}
        \label{fig:M2}
        \includegraphics[width=\textwidth]{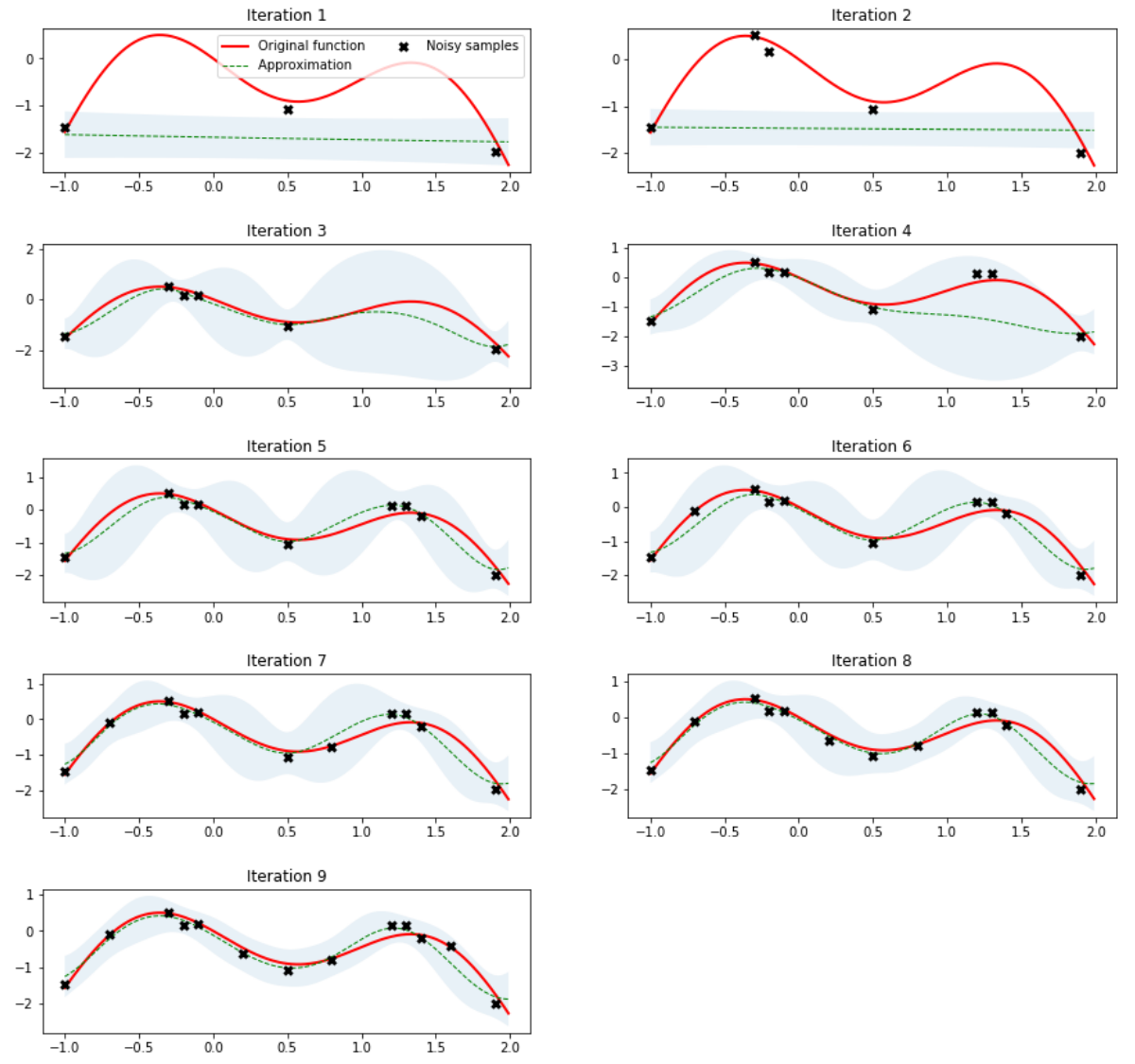}
    \end{subfigure}
\end{adjustwidth}
\caption{Surprise driven experimentation.}    \label{fig:Total}
\end{figure*}

The iteration by iteration approximation performance following the surprise-reacting experiment policy are shown in Fig.~\ref{fig:Total}, in which the left panel, i.e., Fig.~\ref{fig:M1}, presents the approximation iteration based on the Shannon surprise, whereas the right panel, i.e., Fig.~\ref{fig:M2}, presents the approximation iteration based on the Bayesian surprise. The threshold parameter in the Shannon surprise is chosen as $k_\text{Shannon} =1.96$ and that for the Bayesain suprise is chosen to be $k_\text{Bayesian} =0.5$. By using $k_\text{Bayesian} =0.5$, the Bayesian surprise is to flag a distribution change of the magnitude of approximately one standard deviation, which is smaller than that to be flagged in the Shannon surprise.  This small magnitude is used to compensate the slowness of the Bayesian surprise.

If we look at Fig.~\ref{fig:M1}, at first, using the two initial experiment locations ($x=-1.0$ and $x=2.0$) and their corresponding responses, the statistical model does not have an accurate understanding of the underlying function. Rather its model output is a flat response line as shown in iteration 1. The shaded region is the 95\% credible band. Then the sequential algorithm chooses one random location and as the response behaves significantly differently than what the current statistical model anticipates,  the sequential algorithm is presumably surprised after assessing the degree of surprise. So, as a reaction the sequential algorithm then selects a nearby location and do an experiment there as the next sample. The sequential algorithm is surprised again, which confirms that the surprise observation is a valid response and they are both used to update the statistical model so that the updated response starts to move away from the flat line and adapt to the underlying function. This whole surprise confirming action is done in a single iteration.

In iteration 2, after the update, to gain more knowledge about the local region it does a follow-up experiment nearby again. However, it will not be surprised anymore and following the proposed policy, the sequential algorithm will go back to exploration and selects a distant sample in iteration 3, at which location, the sequential algorithm is surprised again. The subsequent exploitation confirms again the validity of the response and further adopts the response closer to the underlying true function.  This process will continue until the experimentation budget is reached at iteration 9 (consuming 13 experiments). Bayesian surprise measure also worked in a similar manner in Fig.~\ref{fig:M2}.  Both surprise metrics are able to reach a good approximation of the underlying function at the end of the experiment (experiment 13).


Using Shannon surprise, the number of experiments consumed is \{2, 4, 5, 7, 8, 9, 10, 11, 12, 13\}, whereas using Bayesian surprise, the number sequence is \{2, 3, 5, 6, 8, 9, 10, 11, 12, 13\}. It is apparent that the difference between the two surprise measures takes place in the earlier iterations. If we look at the first two iterations of these two surprise measures, we can understand the difference between two measures. Shannon surprise, being more sensitive, declares the middle point ($x=0.5$, first sequential experiment output) as surprise and carry out confirmatory experiments near this location. On the contrary, Bayesian surprise declares this event as normal and move to carry out exploration in other regions. Using both approaches reaches a reasonable approximation of the underlying function at the end of iteration 6 and after that, there are not much differences using either surprise measure.


We want to highlight that using the surprise-reacting policy, we successfully identified both the left and right peaks and the valley in between. This is one of the benefits as we consider the surprise-reacting policy, i.e., they may not be the quickest for honing in on the exact optima but they are good for design space or response surface approximation.

\subsubsection{EI-based BO experiment policy}
As a representation of the existing BO acquisition functions, let us now discuss the EI-based BO experiment policy and see how the sequential learning policy behaves differently. This BO experiment policy uses the popular EI acquisition function.  Bayesian optimization tries to find the optimum of inputs, $\mathbf {x_{\text{opt}}}$, so that we can attain the global maximum (or minimum)~\cite{talapatra:2018} of the design space, i.e.,
\begin{equation}
\label{eq:BO}
\mathbf{x_{\text{opt}}}= \argmax_{\mathbf {x} \in \mathbb{R}^{d}} f(\mathbf{x}).
\end{equation}

The acquisition function is the key in a BO framework for deciding where to conduct the next experiment. The EI acquisition function is expressed as~\cite{jones:1998}:
\begin{equation}
\label{eq:EI}
\text{EI}_n(\boldsymbol{x})=(\mu_{*}(\boldsymbol{x})-f(\boldsymbol{x}^{+}))\Phi \left(\frac{\Delta _{n}(\boldsymbol{x})}{\sigma_{*}(\boldsymbol{x})}\right) + \sigma_{*}(\boldsymbol{x})\phi\left(\frac{\Delta _{n}(\boldsymbol{x})}{\sigma_{*}(\boldsymbol{x})}\right),
\end{equation}
where $\Delta _{n}(\boldsymbol{x})=\mu_{*}(\boldsymbol{x})-f(\boldsymbol{x}^{+})$  and it captures the potential improvement over the current best solution ($\boldsymbol{x}^{+}$) if the sequential algorithm chooses $\boldsymbol{x}$ as the new experiment location, $\mu_{*}(\boldsymbol{x})$ and $\sigma_{*}(\boldsymbol{x})$ represent the mean and standard deviation of the GP posterior predictive at $\boldsymbol{x}$, and $\Phi$ and $\phi$ are the cdf and pdf of the standard normal distribution, respectively. The first component of Equation \eqref{eq:EI} favors exploitation and the second component favors exploration. A high value of improvement over the current best solution (component 1) and a high uncertainty (component 2) both result in a final high EI. Equation \eqref{eq:EI} decides the trade-off between the two considerations.  To apply the EI-based experiment policy to the same example explained in Section~\ref{sec3.5.1}, all settings are kept the same as in the surprise-based policy, including the initial experiments and the GP-based statistical model. The key difference is the mechanism of exploitation-exploration switching and where to select the next experiment.

\begin{figure*}[t]
\centering
\includegraphics[scale=0.65]{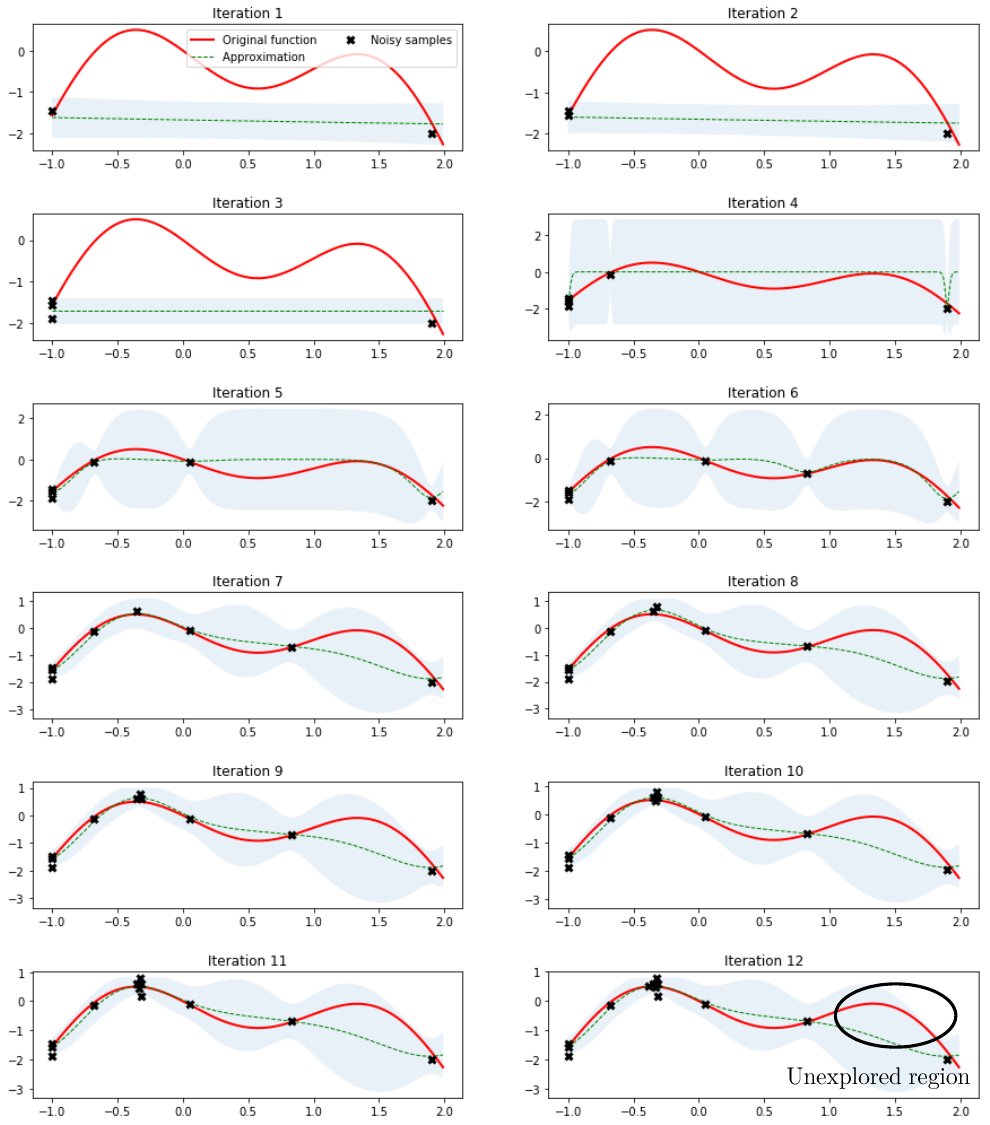}
\caption{EI-based experimentation policy.} \label{Figure5}
\end{figure*}

The experiment process using the EI-based policy is illustrated in Fig.~\ref{Figure5}. One can find that BO is able to locate the maximum value of the underlying function successfully but in that process, the EI-based policy fails to approximate the right half of the functional space. Such an outcome is much expected considering the design of the EI acquisition function itself.  If one looks at the construction of Equation \eqref{eq:EI}, we notice that potential improvement ($\Delta _{n}$) of the candidate locations are weighted by their respective variances. Moreover, the cdf of the variance-corrected improvements is multiplied by the actual improvement to form the exploitation component, while the corresponding pdf is multiplied by the variance of the candidate location to form the exploration component. As such, the first component of Equation \eqref{eq:EI} often overweighs the second component, so much so that unless the improvement is very minimum in regions with low variability, EI hesitates to move towards a high-reward-high-variability region.

In summary, the EI-based policy is in favor of locations that provide a small improvement with more certainty (low $\sigma_{*}(\boldsymbol{x}$)) over a bigger improvement with less certainty (high $\sigma_{*}(\boldsymbol{x}$)). Such policy tends to over-exploit local peaks and could be trapped in local optima. If we look at iteration 12 in the EI policy, we find that the sequential algorithm is stuck to the neighborhood of the left peak. In order for the EI policy to explore the right peak, it could take many more samples.

One may argue that the objective of the EI-based policy is not to fully explore a design space or a response surface. Rather it is to hone in on optima rapidly.  We agree that the EI-based policy is doing a fair job for honing in the optima.  But we want to stress the importance of characterizing response surfaces or design spaces, a well-established need shared across multiple disciplines such as biological systems~\cite{lovell:2011}, energy field~\cite{makela:2017}, or machining process~\cite{habib:2009}.  We hope to convey the message that the surprise-reacting policy provides a better alternative than the EO-based policy.

\section{Performance Comparison }\label{sec:performance}
We here present the performance comparison between the proposed surprise-reacting approach and several well-established acquisition functions utilized in a BO framework for the task of function approximation. When choosing acquisition functions, we prioritize those that are typically used in sequential experimentation on an autonomous platform \cite{burger:2020, liang:2021}. These include EI \cite{jones:1998}, UCB \cite{srinivas:2010}, PI \cite{kushner:1964}, and MaxVar \cite{mockus:1978}.

These acquisition functions are chosen because they represent different strategies in balancing exploration and exploitation.  They employ different criteria to select the experimental locations. EI selects points that maximize the expected improvement over the current best observation, tending to favor exploitation. UCB balances exploration and exploitation by considering both the mean and variance of the prediction, thus encouraging exploration. PI selects points that maximize the probability of improvement over the current best observation. MaxVar selects points with the highest prediction variance, targeting regions of high uncertainty and favoring exploration.

We employ five synthetic benchmark functions and a real-life dataset to test the efficacy of the competing approaches. For a fair comparison, we use the same initial experiments for all the competing approaches. A GP-based statistical model will be used for all the competing approaching including the surprise-reacting policy under the same resource-constrained environment. To compare the prediction performance of the underlying response surface, we use both the Root Mean squared Error (RMSE) and the Continuous Ranked Probability Score (CRPS) as the performance metrics. The RMSE is defined as:
\begin{equation}
\label{eq:RMSE}
\text{RMSE}=\sqrt{\frac{1}{T}\sum_{j=1}^{j=T}(y_{j}-\widehat{y}_{j})^{2}},
\end{equation}
where \( T \) is the size of the test set, \( \widehat{y}_{j} \) is the prediction made by the trained statistical model, and \( y_{j} \) is the true response. For the simulated datasets, \( y_{j} \) is generated by the underlying true function, whereas in the real-life dataset, \( y_{j} \) is the actual physical measurement. A lower RMSE corresponds to better function approximation.

The CRPS is defined as:
\begin{equation}
\label{eq:CRPS}
\text{CRPS} = \frac{1}{T} \sum_{j=1}^{T} \int_{-\infty}^{\infty} \left( F_{j}(z) - \mathbf{1}\{z \geq y_{j}\} \right)^{2} dz,
\end{equation}
where \( F_{j} \) is the cumulative distribution function (CDF) of the predictive distribution for \( y_{j} \), and \( z \) is a variable over which the CDF is integrated. The CRPS measures the difference between the predicted cumulative distribution and the actual value, with a lower CRPS indicating better probabilistic predictions.

\begin{table*}[h]
    \centering
    \caption{Performance comparison of different acquisition functions using RMSE (mean $\pm$ 95\% CI) values. Best performance in each column is highlighted in bold.}
    \begin{tabular}{lcccccc}
        \hline
        & \textbf{Branin} & \textbf{Six-Hump Camelback} & \textbf{Ackley} & \textbf{Rastrigin} & \textbf{Sum Squares} \\
        \textbf{Acquisition Functions} &  &  &  &  &  \\
        \hline
        Shannon Surprise & $\mathbf{0.75 \pm 0.09}$ & $\mathbf{2.67 \pm 0.19}$ & $\mathbf{0.80 \pm 0.07}$ & $\mathbf{28.79 \pm 1.03}$ & $165.80 \pm 9.77$ \\
        Bayesian Surprise & $1.16 \pm 0.12$ & $2.73 \pm 0.20$ & $0.84 \pm 0.07$ & $36.50 \pm 1.38$ & $\mathbf{150.09 \pm 6.54}$ \\
        EI & $9.18 \pm 4.97$ & $10.42 \pm 0.88$ & $4.95 \pm 2.22$ & $41.19 \pm 8.17$ & $432.87 \pm 70.60$ \\
        PI & $9.14 \pm 1.08$ & $15.17 \pm 1.24$ & $5.70 \pm 2.29$ & $31.31 \pm 2.99$ & $223.91 \pm 15.27$ \\
        UCB & $4.75 \pm 3.09$ & $5.71 \pm 0.60$ & $4.48 \pm 2.16$ & $31.66 \pm 5.17$ & $378.41 \pm 65.17$ \\
        MaxVar & $12.57 \pm 7.14$ & $2.85 \pm 0.12$ & $5.33 \pm 2.33$ & $36.53 \pm 6.53$ & $389.02 \pm 70.68$ \\
        \hline
    \end{tabular}
    \label{tbl:p1}
\end{table*}

\begin{table*}[h]
    \centering
    \caption{Performance comparison of different acquisition functions using CRPS (mean $\pm$ 95\% CI) values. Best performance in each column is highlighted in bold.}
    \begin{tabular}{lcccccc}
        \hline
        & \textbf{Branin} & \textbf{Six-Hump Camelback} & \textbf{Ackley} & \textbf{Rastrigin} & \textbf{Sum Squares} \\
        \textbf{Acquisition Functions}  \\
        \hline
        Shannon Surprise & $\mathbf{0.64 \pm 0.02}$ & $\mathbf{1.51 \pm 0.07}$ & $\mathbf{0.47 \pm 0.04}$ & $\mathbf{17.05 \pm 0.55}$ & $97.87 \pm 6.80$ \\
        Bayesian Surprise & $0.93 \pm 0.03$ & $1.69 \pm 0.06$ & $0.49 \pm 0.04$ & $23.20 \pm 0.58$ & $\mathbf{85.50 \pm 4.09}$ \\
        EI & $4.27 \pm 2.82$ & $4.75 \pm 0.47$ & $2.93 \pm 1.34$ & $24.52 \pm 4.88$ & $287.92 \pm 53.66$ \\
        PI & $3.73 \pm 0.42$ & $7.35 \pm 0.69$ & $3.40 \pm 1.39$ & $19.85 \pm 1.62$ & $129.26 \pm 10.02$ \\
        UCB & $2.32 \pm 1.77$ & $2.38 \pm 0.19$ & $2.66 \pm 1.31$ & $18.53 \pm 3.11$ & $245.14 \pm 49.42$ \\
        MaxVar & $7.39 \pm 4.07$ & $1.57 \pm 0.05$ & $3.17 \pm 1.40$ & $21.80 \pm 3.85$ & $260.13 \pm 53.40$ \\
        \hline
    \end{tabular}
    \label{tbl:p2}
\end{table*}

\subsection{Synthetic Benchmark Functions}
We consider several popular benchmark functions routinely used for testing the performance of BO-based acquisition functions and global optimization algorithms~\cite{de2015, lee2020}. Among these are the six-hump camelback function and the Branin function, both of which are two-dimensional. The six-hump camelback function is typically evaluated within bounds of \([-3, 3]\) for the first dimension and \([-2, 2]\) for the second dimension, and the Branin function within bounds of \([-5, 10]\) for the first dimension and \([0, 15]\) for the second dimension. Additionally, we utilize higher-dimensional functions such as the Ackley, Rastrigin, and Sum Squares functions. The Ackley function is evaluated within bounds of \([-32.768, 32.768]\) in five dimensions, the Rastrigin function within \([-5.12, 5.12]\) in five dimensions, and the Sum Squares function within \([-10, 10]\) in six dimensions. 

We allocate 10 initial experiments and 50 sequential experiments to the competing approaches, making a total of 60 experiments to learn the function. While allowing more evaluations would likely result in all approaches approximating the function well in the long run, our goal is to assess how well they approximate with a fixed budget. After completing these experiments, the trained statistical models from all competing approaches are used to perform predictions ($\widehat{y}_{j}$) on an independent test set of 50 locations, i.e., $T=50$. The RMSEs and CRPSs are then calculated for each approach. This process is repeated 50 times to quantify the uncertainty in RMSE and CRPS.

Tables \ref{tbl:p1} and \ref{tbl:p2} summarize the performance comparison among the competing approaches for the five benchmark functions, using RMSE and CRPS metrics, respectively. For both RMSE and CRPS, we report the mean $\pm$ 95\% Confidence Interval (CI). The results indicate that the surprise-reacting approaches achieve superior prediction performance, with Shannon surprise outperforming Bayesian surprise in most cases. Additionally, the prediction variability is much lower for most of the cases when using surprise measures compared to other competing methods.

EI performs poorly compared to the other approaches, reflecting its tendency to over-exploit the current best observations rather than exploring the function space adequately. PI, while slightly better than EI, still underperforms due to a similar tendency to favor exploitation. UCB shows better performance than the other BO acquisition functions in general, as it balances exploration and exploitation by considering both the mean and variance of the predictions. However, it still lags behind the surprise-based approaches. MaxVar, which focuses on selecting points with the highest prediction variance, performs inconsistently across different functions, indicating that while it encourages exploration, it may not effectively balance it with exploitation.

\begin{figure}[H]
    \centering
    \begin{subfigure}[b]{0.48\textwidth}
        \includegraphics[width=\linewidth]{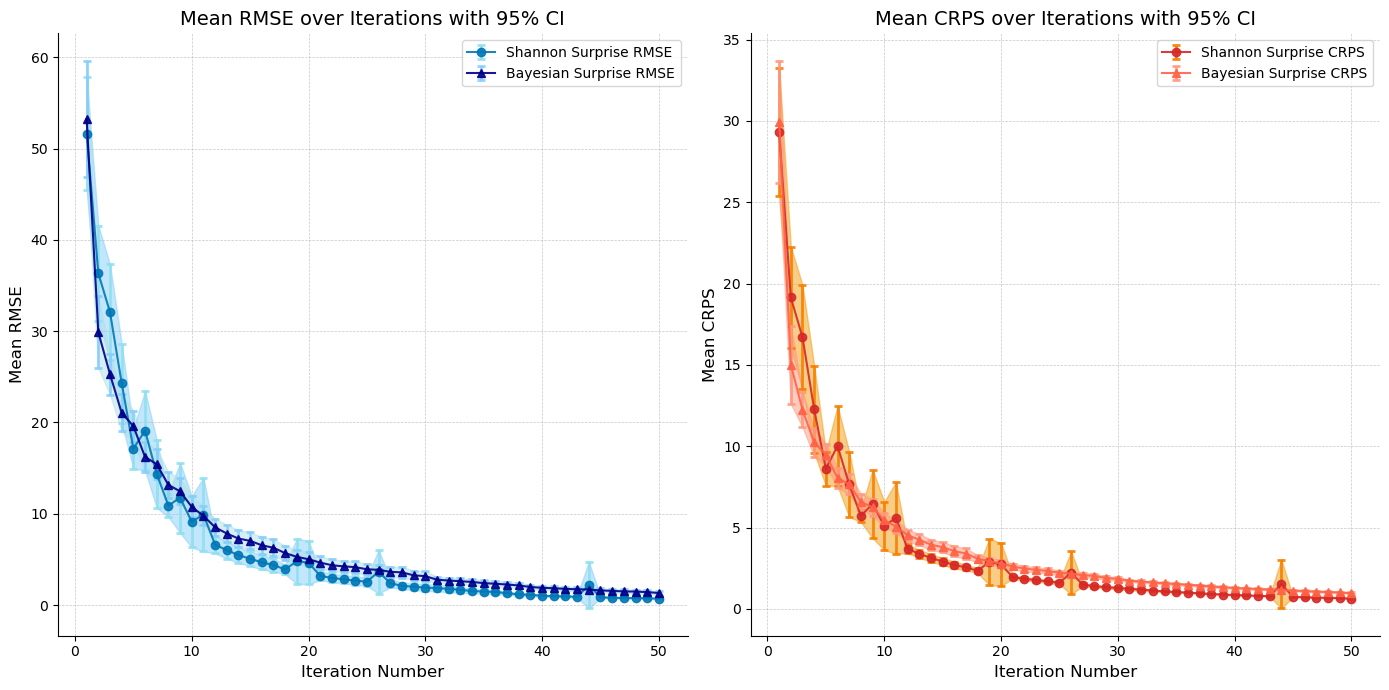}
        \caption{Performance of Surprise Measures}
        \label{fig:figure1}
    \end{subfigure}
    \vfill
    \begin{subfigure}[b]{0.48\textwidth}
        \includegraphics[width=\linewidth]{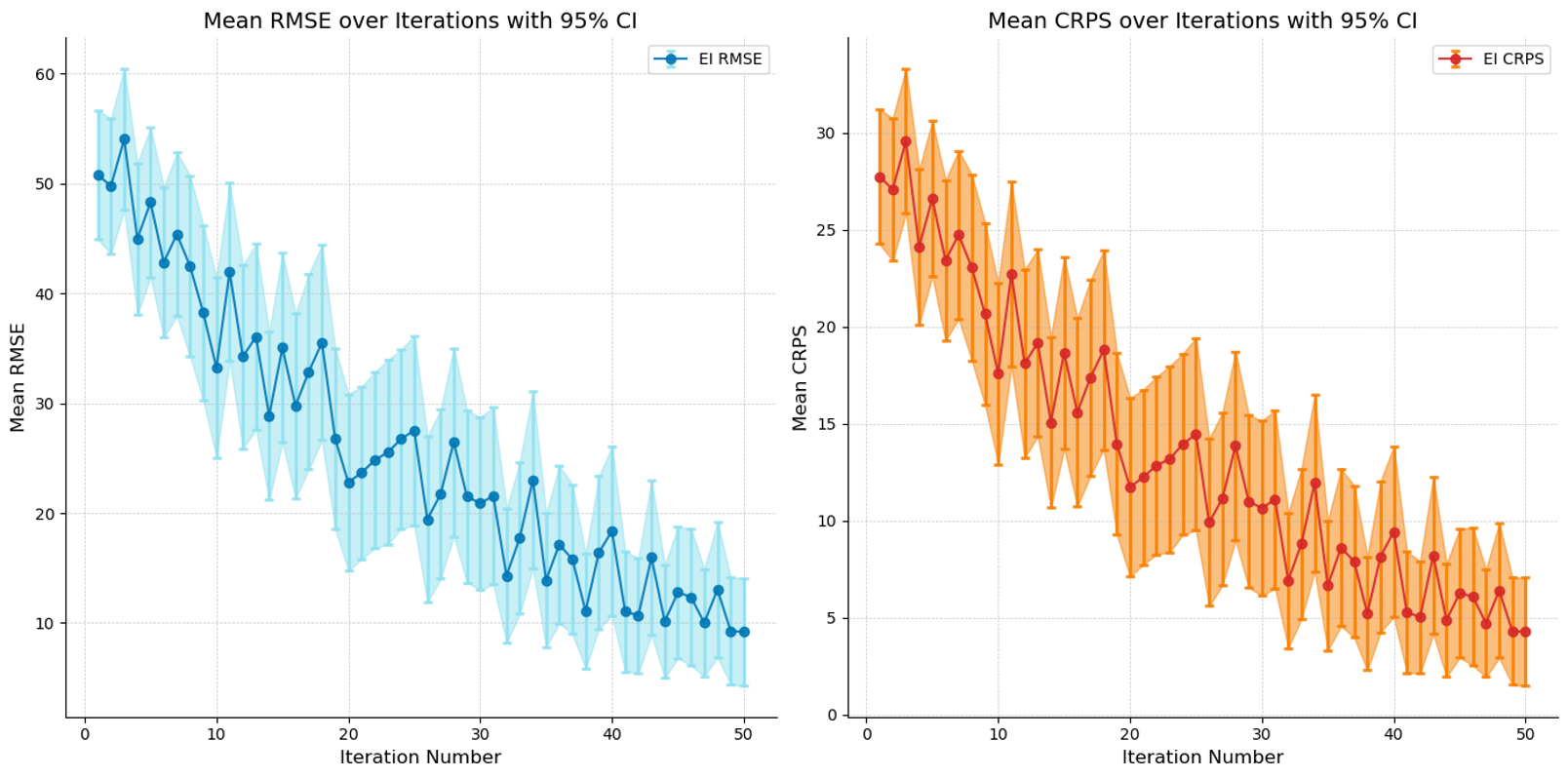}
        \caption{Performance of EI}
        \label{fig:figure2}
    \end{subfigure}
    \vfill
    \begin{subfigure}[b]{0.48\textwidth}
        \includegraphics[width=\linewidth]{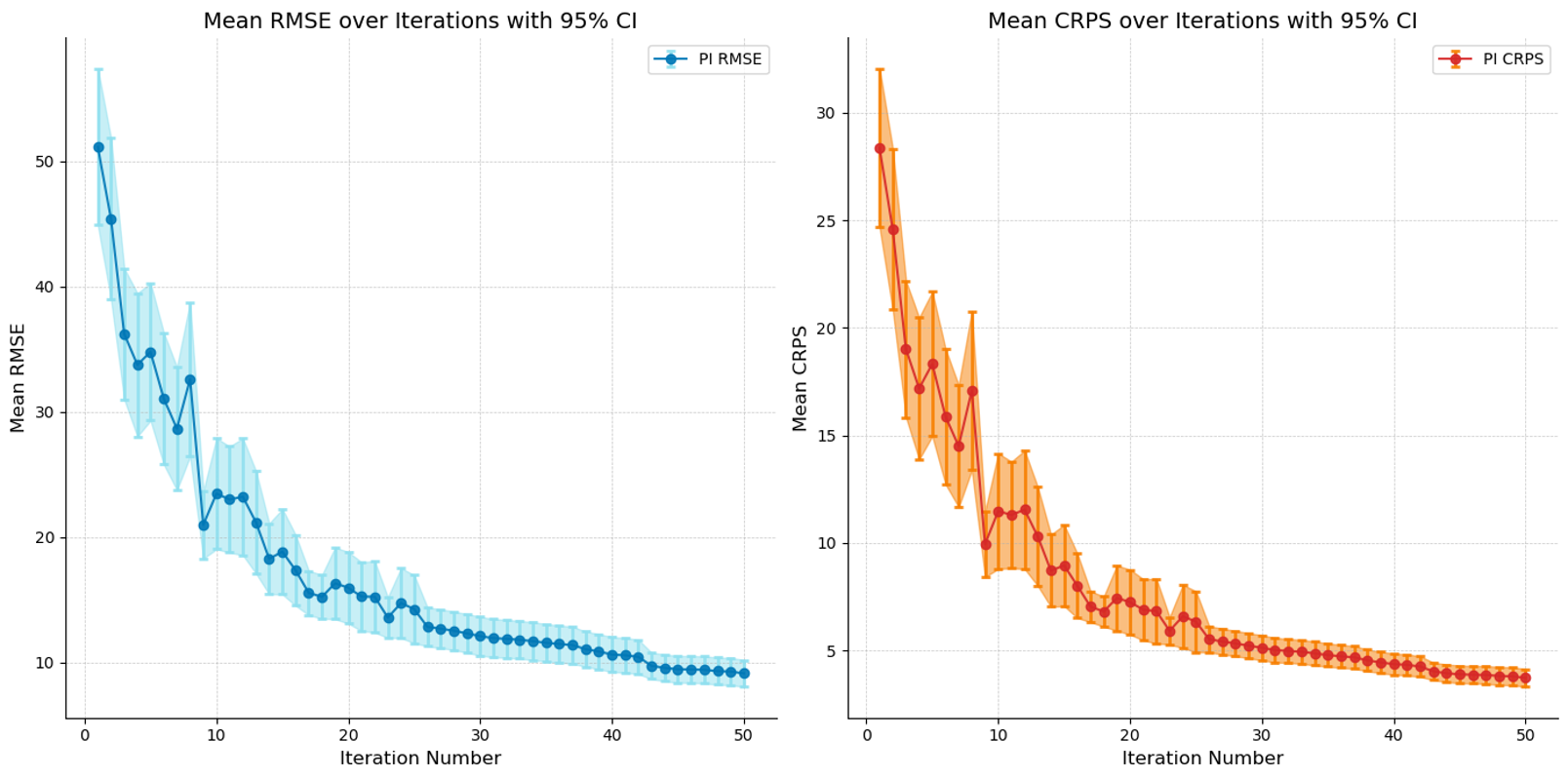}
        \caption{Performance of PI}
        \label{fig:figure3}
    \end{subfigure}
    \vfill
    \begin{subfigure}[b]{0.48\textwidth}
        \includegraphics[width=\linewidth]{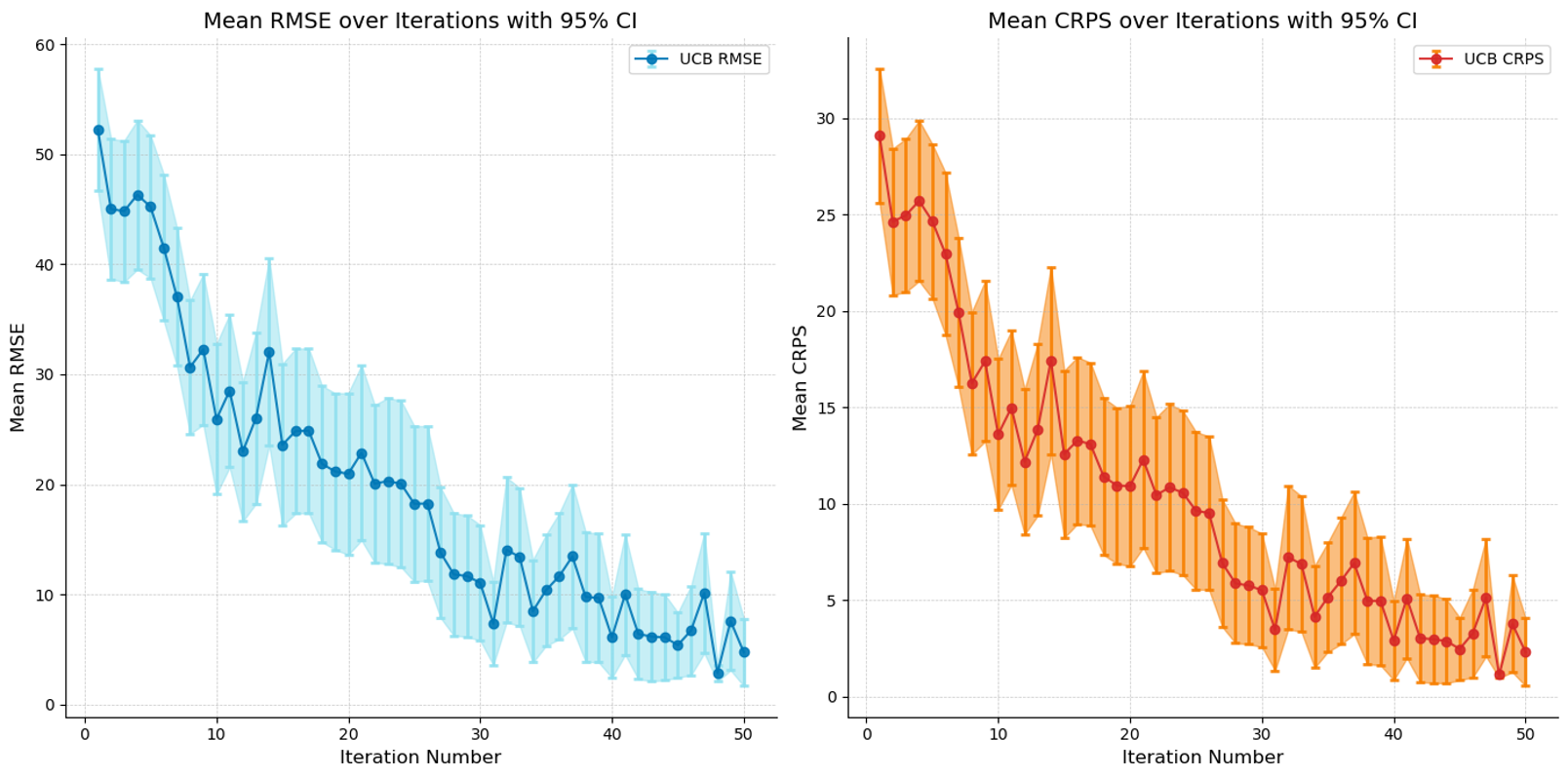}
        \caption{Performance of UCB}
        \label{fig:figure4}
    \end{subfigure}
    \vfill
    \begin{subfigure}[b]{0.48\textwidth}
        \includegraphics[width=\linewidth]{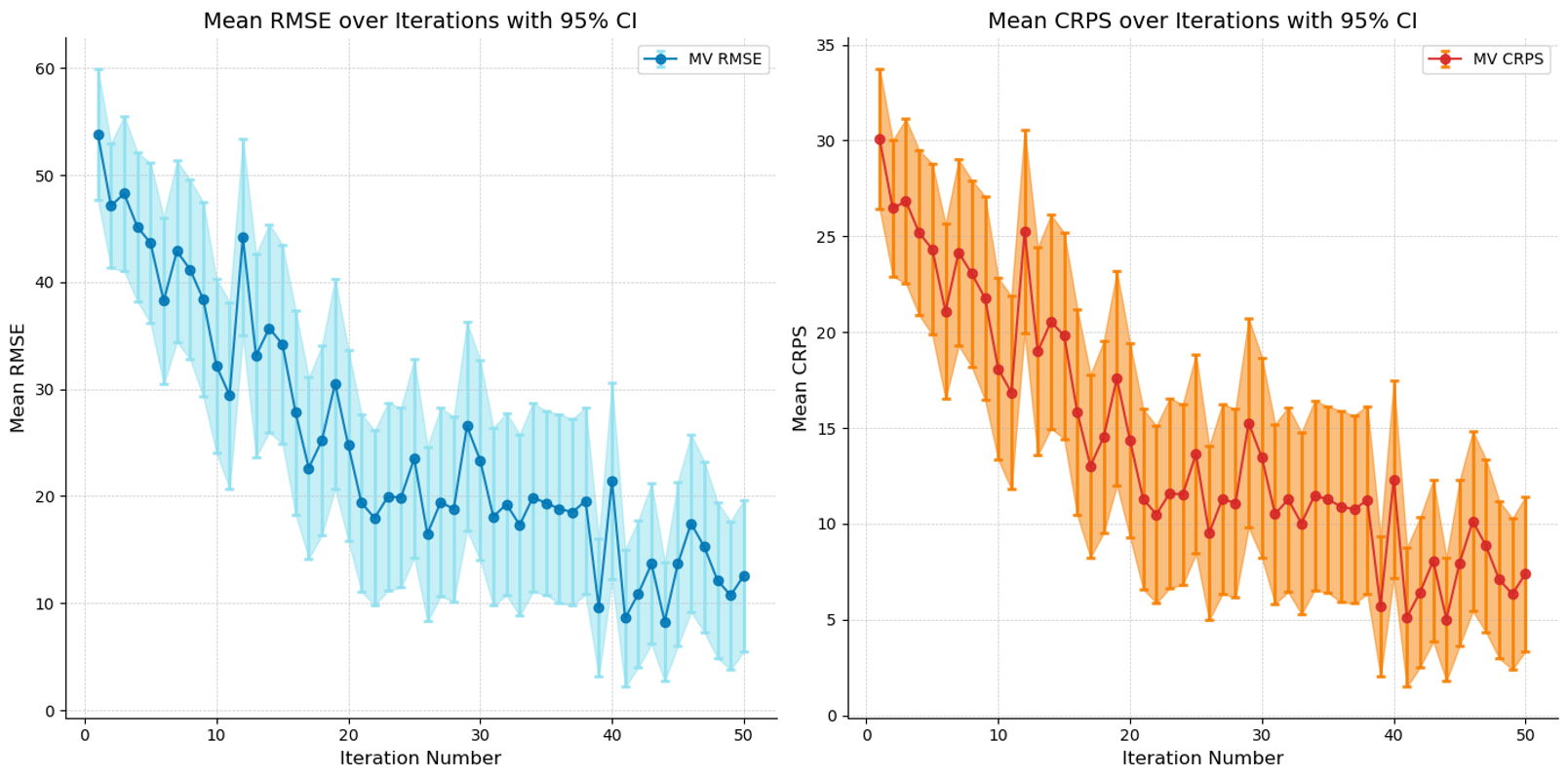}
        \caption{Performance of MaxVar}
        \label{fig:figure5}
    \end{subfigure}
    \caption{Iterative approximation performance for the Branin function}
    \label{fig:all_figures}
\end{figure}

\begin{figure}[H]
    \centering
    \begin{subfigure}[b]{0.48\textwidth}
        \includegraphics[width=\linewidth]{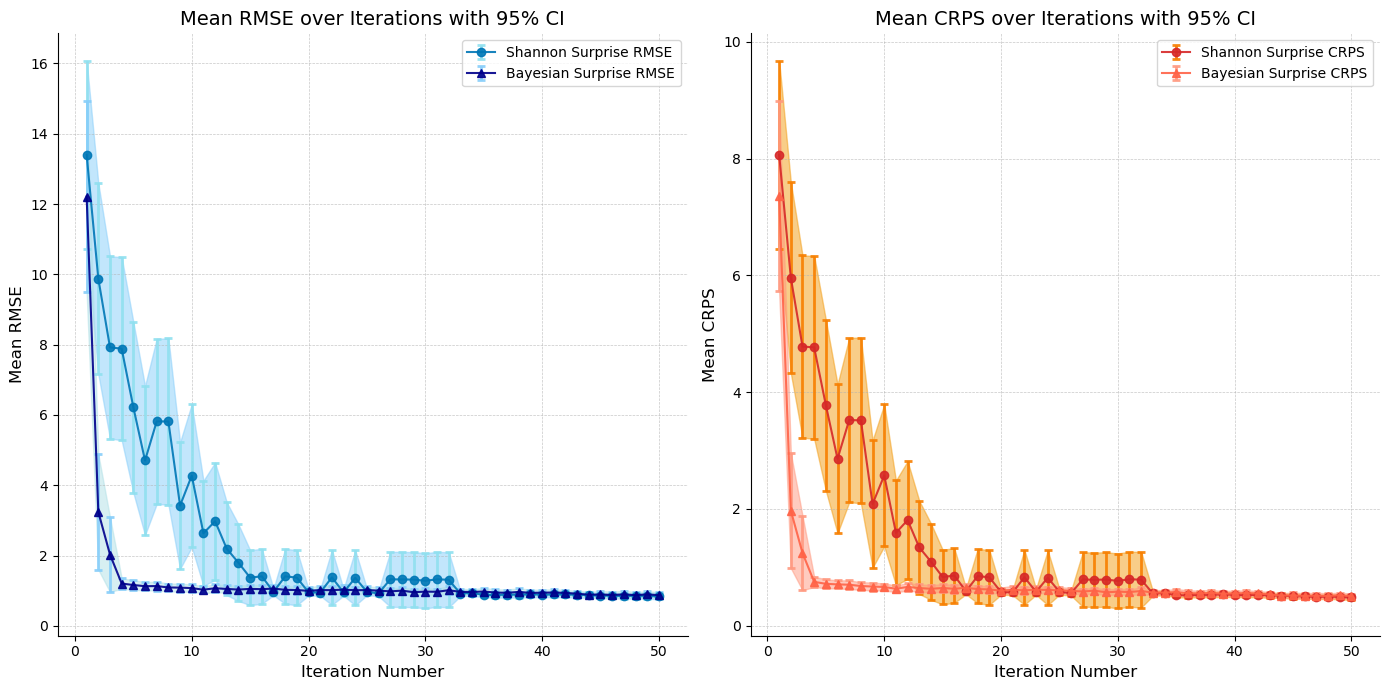}
        \caption{Performance of Surprise Measures}
        \label{fig:figure11}
    \end{subfigure}
    \vfill
    \begin{subfigure}[b]{0.48\textwidth}
        \includegraphics[width=\linewidth]{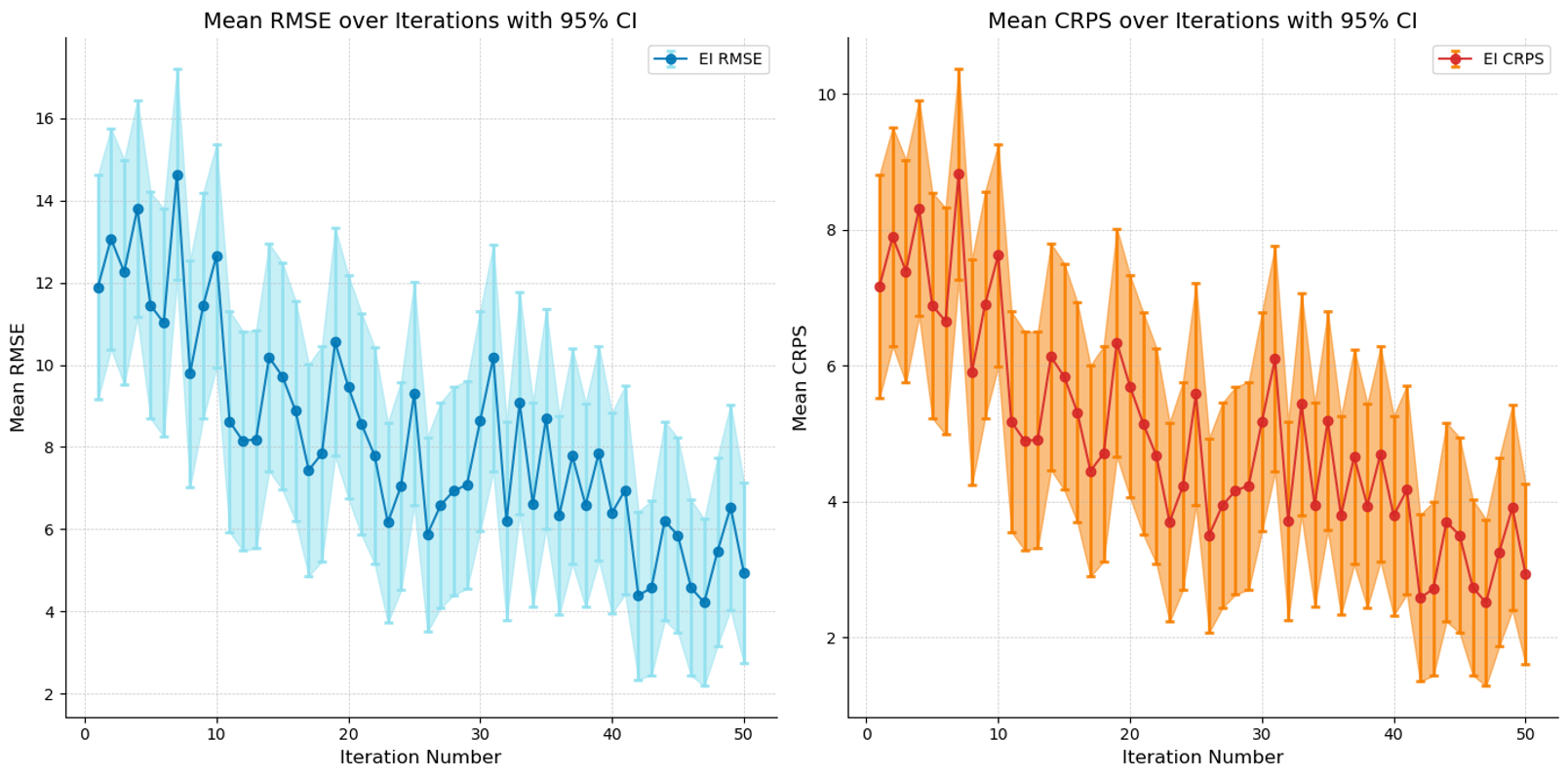}
        \caption{Performance of EI}
        \label{fig:figure12}
    \end{subfigure}
    \vfill
    \begin{subfigure}[b]{0.48\textwidth}
        \includegraphics[width=\linewidth]{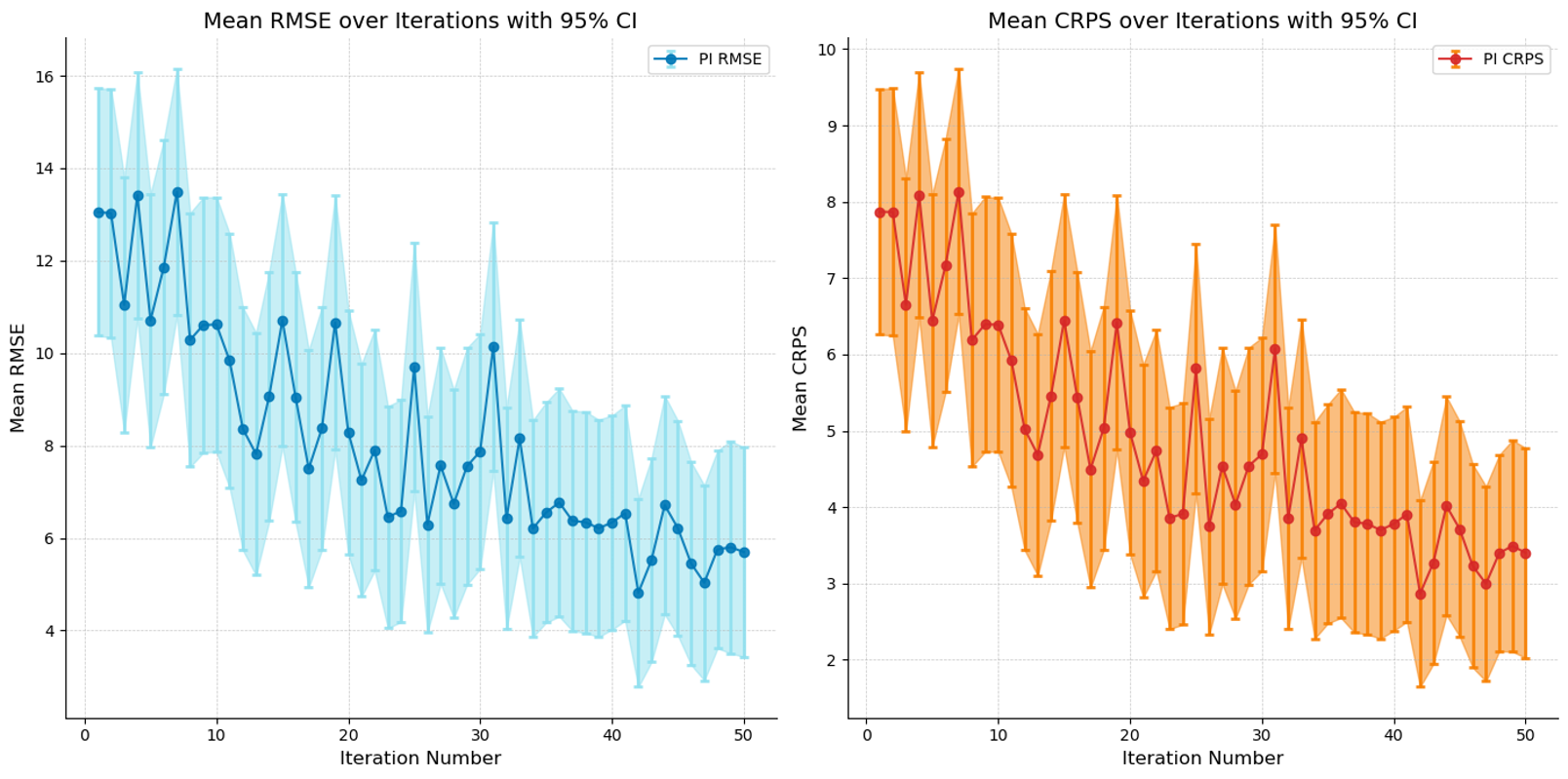}
        \caption{Performance of PI}
        \label{fig:figure13}
    \end{subfigure}
    \vfill
    \begin{subfigure}[b]{0.48\textwidth}
        \includegraphics[width=\linewidth]{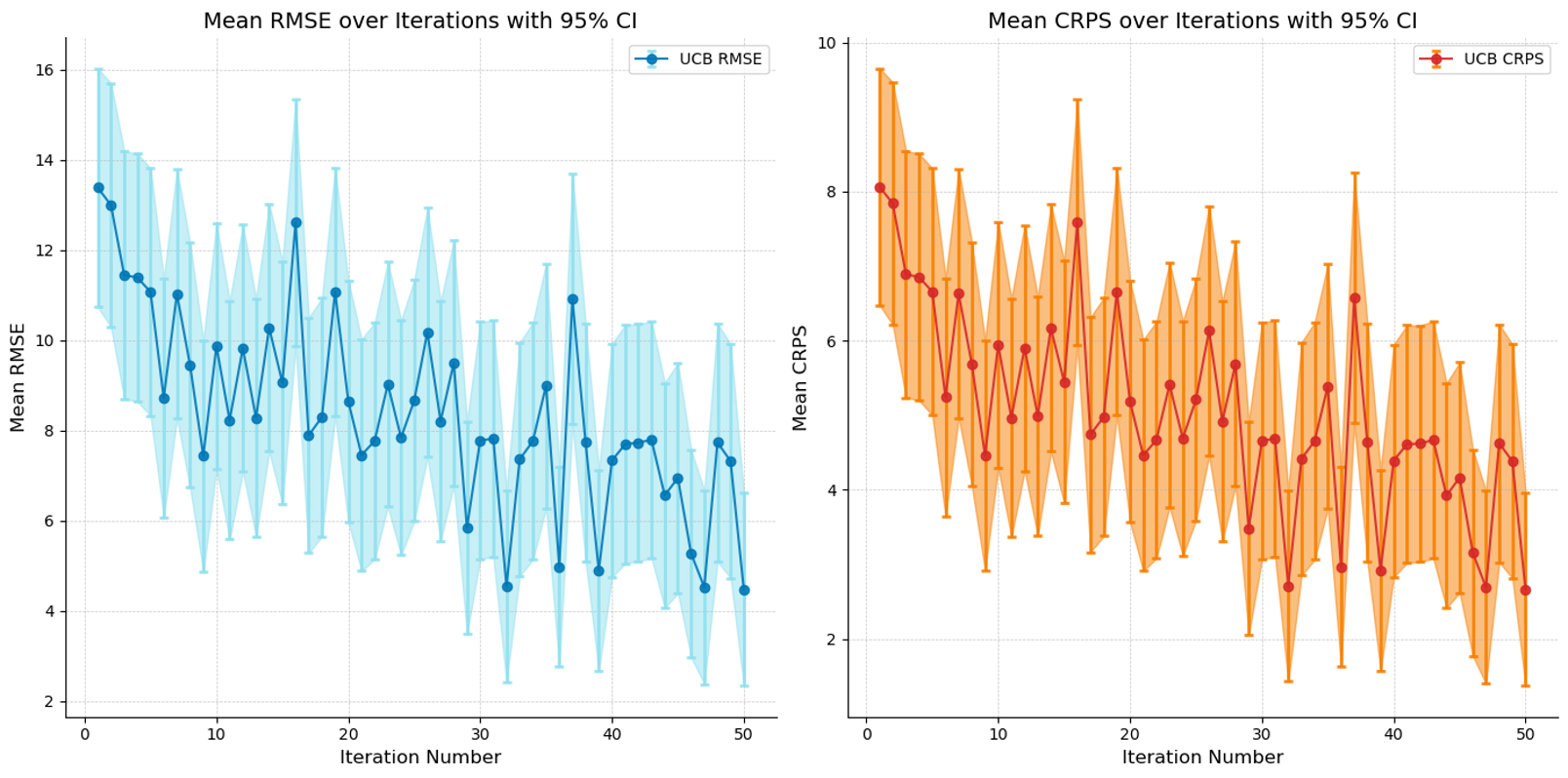}
        \caption{Performance of UCB}
        \label{fig:figure14}
    \end{subfigure}
    \vfill
    \begin{subfigure}[b]{0.48\textwidth}
        \includegraphics[width=\linewidth]{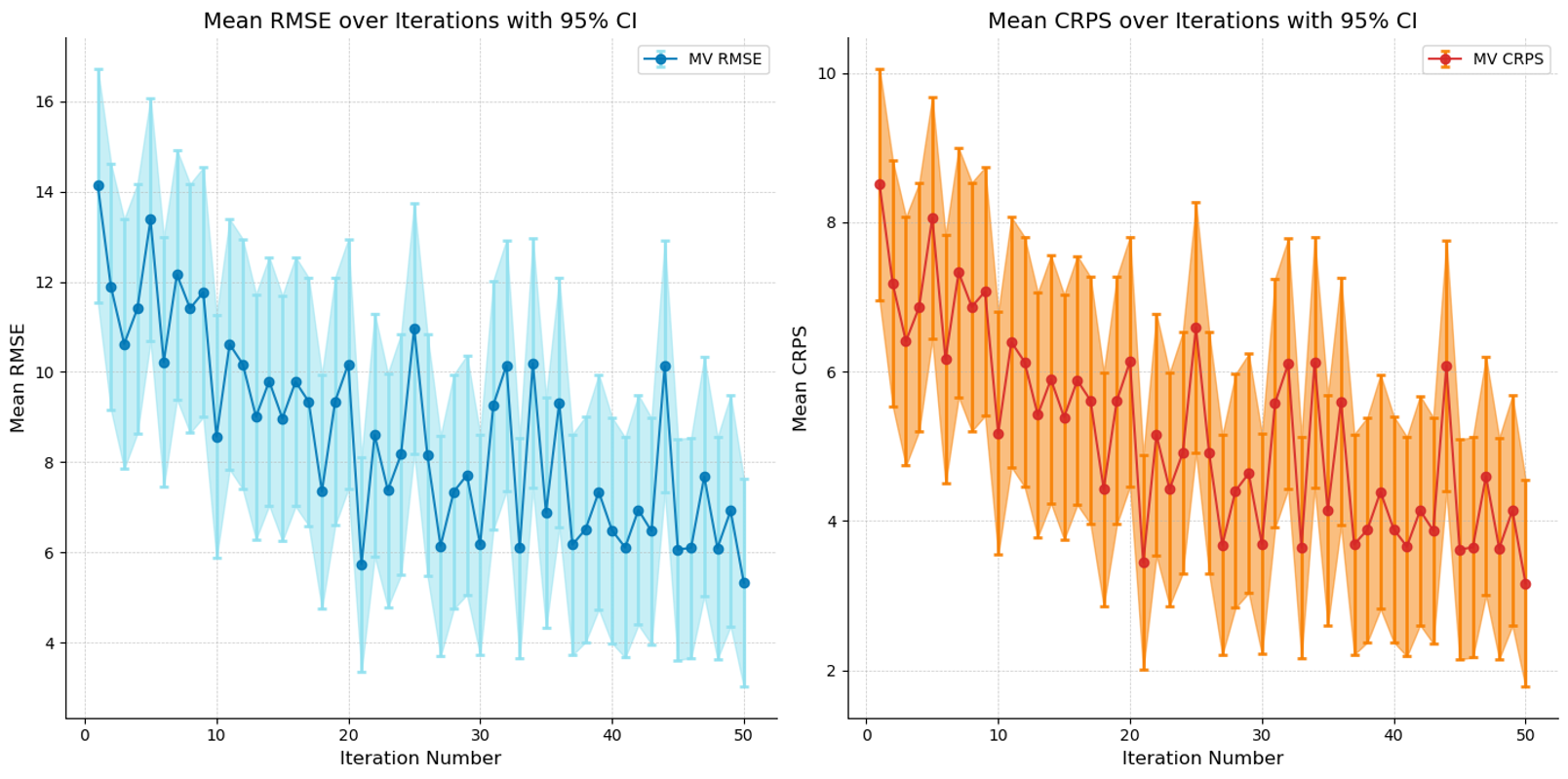}
        \caption{Performance of MaxVar}
        \label{fig:figure15}
    \end{subfigure}
    \caption{Iterative approximation performance for the Ackley function}
    \label{fig:ack}
\end{figure}

Additionally, to evaluate the performance of the competing approaches as the sequential experiment progresses, we record their performance at each iteration until the budget of 50 experiments is reached. When plotting these iteration-wise RMSE and CRPS values for all the benchmark functions, we observe that the surprise-reacting approaches achieve a rapid decrease in both RMSE and CRPS values compared to the other acquisition functions. Due to space limitations, we only present these plots for the Branin and Ackley functions in Fig.~\ref{fig:all_figures} and Fig.~\ref{fig:ack}, respectively.  But the same message holds for other benchmarking functions.

Overall, the results suggest that surprise-reacting approaches, particularly those using Shannon surprise, provide a robust method for function approximation under limited evaluation budgets, effectively balancing the trade-off between exploration and exploitation.

\subsection{Impact of Threshold}

In the evaluation presented in the previous subsection, we compute the RMSE and CRPS for both Shannon and Bayesian surprise measures using specific threshold parameters to determine surprise events. For Shannon surprise, \( k_\text{Shannon} = 1.96 \) is used to signify a 95\% credible interval, arguably the most commonly used C.I.. For Bayesian surprise, \( k_\text{Bayesian} = 0.5 \) corresponding to a less than 95\% C.I. to account for its slow responsiveness.

To assess the impact of varying these threshold parameters, we adjust these values to represent different confidence levels for Shannon surprise: 90\% (\( k_\text{Shannon} = 1.645 \)), 97.5\% (\( k_\text{Shannon} = 2.241 \)), and 99\% (\( k_\text{Shannon} = 2.576 \)). Following a similar approach, the Bayesian surprise threshold parameter is adjusted to 0.42, 0.57, and 0.66, respectively.

The results, presented in Tables~\ref{tbl:threshold_impact} and \ref{tbl:threshold_impact_crps}, indicate that changing the threshold does change the prediction performance of the surprise-reacting policy.  But within a reasonable range of the thresholds (for Shannon, 90\% or above), the difference in performance is acceptable.  There is one case in Bayesian surprise where the performance deviates from the baseline by as much as 15.52\%.  But for the majority of the cases, the performance change is less than 6\%. This suggests that the effect of threshold is manageable, highlighting the degree of robustness of both Shannon and Bayesian surprise measures in approximating the underlying response surface.

\subsection{Grinding Dataset}
In addition to synthetic benchmark functions we also evaluate the performance of our approach on a real-life grinding data set~\cite{botcha:2021}. This is not the first time that this surprise-reacting policy was used on a real system.  As mentioned in Introduction, Jin et al. \cite{jin:2022} applied an earlier version of the surprise-reacting policy to a 3D printing process and reported that for achieving the same quality output, the surprise-reacting policy used one-third to one-half of the experimental runs that the EI-based approach used. Here we report a second case in which an improvement is confirmed.

For the grinding process~\cite{sanjeevi:2021}, the purpose is to quickly characterizing the response surfaces using a limited amount of resources, so that the operators can get the desired surface roughness in different stages of the surface-finishing process by setting the correct process conditions, such as the feed rate and speed.  The dataset is from a cylindrical plunge grinding process. Three process parameters ($\mathbf x$) are the work speed, wheel speed, and the in-feed of the grinding wheel. The goal is to establish a relationship between these process parameters and the surface roughness. The actual surface roughness is measured at the end of each experiment and the physical measurement is treated as $y$, and the $\widehat{y}$ is obtained from the trained statistical model.

\begin{table*}[h]
    \centering
    \caption{Impact of the threshold change on the performance of Shannon surprise and Bayesian surprise measures using RMSE values. The percentage change relative to the use of the base threshold is included in the parentheses.}
    \scalebox{0.90}{
    \begin{tabular}{lccc|ccc}
        \toprule
        & \multicolumn{3}{c}{\textbf{Shannon Surprise (Base threshold: 1.96)}} & \multicolumn{3}{c}{\textbf{Bayesian Surprise (Base threshold: 0.5)}} \\
        \cmidrule(lr){2-4} \cmidrule(lr){5-7}
        & \textbf{1.65} & \textbf{2.24} & \textbf{2.58} & \textbf{0.42} & \textbf{0.57} & \textbf{0.66} \\
        \midrule
        \textbf{Branin} & $0.78 \pm 0.14$ (4.00\%) & $0.78 \pm 0.13$ (4.00\%) & $0.74 \pm 0.09$ (-1.33\%) & $1.34 \pm 0.29$ (15.52\%) & $1.19 \pm 0.13$ (2.59\%) & $1.14 \pm 0.14$ (-1.72\%) \\
        \textbf{Hump} & $2.69 \pm 0.20$ (0.75\%) & $2.67 \pm 0.19$ (0.00\%) & $2.83 \pm 0.24$ (5.99\%) & $2.54 \pm 0.20$ (-6.96\%) & $2.91 \pm 0.28$ (6.59\%) & $2.92 \pm 0.32$ (6.96\%) \\
        \textbf{Ackley} & $0.78 \pm 0.06$ (-2.50\%) & $0.79 \pm 0.06$ (-1.25\%) & $0.78 \pm 0.06$ (-2.50\%) & $0.80 \pm 0.07$ (-4.29\%) & $0.83 \pm 0.07$ (-1.25\%) & $0.83 \pm 0.06$ (-1.25\%) \\
        \textbf{Rastrigin} & $28.56 \pm 1.19$ (-0.80\%) & $28.55 \pm 1.27$ (-0.83\%) & $27.99 \pm 0.98$ (-2.78\%) & $35.51 \pm 1.57$ (-2.71\%) & $34.66 \pm 1.39$ (-5.04\%) & $35.59 \pm 1.43$ (-2.49\%) \\
        \textbf{Sum Squares} & $172.13 \pm 8.48$ (3.82\%) & $173.27 \pm 8.88$ (4.51\%) & $174.90 \pm 11.34$ (5.49\%) & $153.56 \pm 7.01$ (2.31\%) & $151.96 \pm 7.04$ (1.25\%) & $146.58 \pm 6.54$ (-2.34\%) \\
        \bottomrule
    \end{tabular}}
    \label{tbl:threshold_impact}
\end{table*}

\begin{table*}[h]
    \centering
    \caption{Impact of the threshold change on the performance of Shannon surprise and Bayesian surprise measures using CRPS values. The percentage change relative to the use of base threshold is included in the parentheses.}
    \scalebox{0.90}{
    \begin{tabular}{lccc|ccc}
        \toprule
        & \multicolumn{3}{c}{\textbf{Shannon Surprise (Base threshold: 1.96)}} & \multicolumn{3}{c}{\textbf{Bayesian Surprise (Base threshold: 0.5)}} \\
        \cmidrule(lr){2-4} \cmidrule(lr){5-7}
         & \textbf{1.65} & \textbf{2.24} & \textbf{2.58} & \textbf{0.42} & \textbf{0.57} & \textbf{0.66} \\
        \midrule
        \textbf{Branin} & $0.63 \pm 0.02$ (-1.56\%) & $0.63 \pm 0.03$ (-1.56\%) & $0.62 \pm 0.02$ (-3.13\%) & $0.96 \pm 0.06$ (3.23\%) & $0.91 \pm 0.03$ (-2.15\%) & $0.92 \pm 0.03$ (-1.08\%) \\
        \textbf{Hump} & $1.51 \pm 0.07$ (0.00\%) & $1.49 \pm 0.06$ (-1.32\%) & $1.52 \pm 0.07$ (0.66\%) & $1.67 \pm 0.07$ (-1.18\%) & $1.71 \pm 0.07$ (1.18\%) & $1.70 \pm 0.08$ (0.59\%) \\
        \textbf{Ackley} & $0.46 \pm 0.03$ (-2.13\%) & $0.46 \pm 0.03$ (-2.13\%) & $0.45 \pm 0.04$ (-4.26\%) & $0.48 \pm 0.04$ (-2.04\%) & $0.48 \pm 0.03$ (-2.04\%) & $0.49 \pm 0.04$ (0.00\%) \\
        \textbf{Rastrigin} & $16.92 \pm 0.58$ (-0.76\%) & $17.08 \pm 0.66$ (0.18\%) & $16.55 \pm 0.59$ (-2.93\%) & $22.72 \pm 0.70$ (-2.07\%) & $22.35 \pm 0.60$ (-3.66\%) & $22.69 \pm 0.67$ (-2.20\%) \\
        \textbf{Sum Squares} & $101.07 \pm 5.36$ (3.27\%) & $102.85 \pm 5.90$ (5.09\%) & $104.12 \pm 7.89$ (6.39\%) & $86.86 \pm 3.83$ (1.59\%) & $86.95 \pm 4.38$ (1.70\%) & $82.41 \pm 3.62$ (-3.61\%) \\
        \bottomrule
    \end{tabular}}
    \label{tbl:threshold_impact_crps}
\end{table*}

\begin{table*}[h]
    \centering
    \caption{Performance comparison of different acquisition functions on Grinding Data using RMSE and CRPS (mean ± 95\% CI). Best performance in each column is highlighted in bold.}
    \begin{tabular}{lcc}
        \hline
        \textbf{Acquisition Function} & \textbf{RMSE} & \textbf{CRPS} \\
   \hline
        Shannon Surprise & $\mathbf{0.157 \pm 0.009}$ & $\mathbf{0.087 \pm 0.004}$ \\
        Bayesian Surprise & $0.165 \pm 0.011$ & $0.101 \pm 0.005$ \\
        EI & $0.165 \pm 0.012$ & $0.091 \pm 0.005$ \\
        PI & $0.162 \pm 0.010$ & $0.091 \pm 0.005$ \\
        UCB & $0.169 \pm 0.012$ & $0.093 \pm 0.006$ \\
        MaxVar & $0.170 \pm 0.011$ & $0.093 \pm 0.005$ \\
        \hline
    \end{tabular}
    \label{tbl:g}
\end{table*}

Each experimentation goes through four stages starting from roughing and then gradually proceeding into semi-finishing, finishing, and then spark out at the very end. Surface roughness measurements are taken at the end of these four stages.  For this grinding process, the experiments have already been carried out before we could apply either the surprise-reacting policy or the competing approaches to it. The existing experiment data are obtained using a central composite design \cite{botcha:2021}. There were a total of 38 experiments, each of which entails four stages. It yielded a total of 152 data samples.

As the experiments involved an expensive Titanium workpiece, it is not easy to repeat it in a brand new experiment. To arrange our test to best reflect a sequential experiment process, we divide the whole dataset of 152 samples into three sets randomly. The first set consists of 10 samples and these are used as initial experiments to train the initial statistical model. A test set of 38 samples have been set aside to test the performance of the competing approaches. The remaining 104 samples provide the possible settings that can be selected in sequential experiments. In other words, as the sequential experimentation policy starts with its process of selecting experiments one at a time, its choices are limited to these 104 samples. We stress that this constraint is equally applied to all the sequential experiment policies.  The experimental budget is fixed at 50 sequential experiments, which makes a total of 60 experiments counting the 10 initial experiments.

Similar to the case of synthetic benchmark functions, we present the comparative performance in Table~\ref{tbl:g}. On this real-life dataset, Shannon surprise performs better compared to Bayesian surprise and the other competing approaches, although the margin of improvement is smaller than that in the benchmark function studies. This is due to the nature of the sequential experiment setup as explained earlier—unlike the benchmark functions, we cannot query an arbitrary candidate location at will but are rather constrained to those already conducted to candidate locations. Overall, we believe that the surprise-reacting policy is able to deliver robust performance in characterizing an unknown, complex response surface quickly.

\subsection{Potential Use in Futuristic Autonomous Platforms}

As outlined at the beginning of the paper, the motivation for our approach arises from its potential application in autonomous platforms to guide the experimentation process. Due to the lack of access to such a platform, we utilized standard benchmark functions and a grinding dataset to simulate real manufacturing experiments. This allows us to evaluate how our method can approximate an unknown response surface within a constrained budget, to some degree of mimicking its expected performance in an autonomous setting. Nevertheless, we acknowledge that there exist several key differences between benchmark functions and real-life scenarios, which may affect the performance of the proposed approach. Below, we outline these differences and their implications:

\begin{enumerate}

\item \textbf{Experimental Uncertainty}: Real-world experiments are subjected to various sources of noise, such as measurement errors and environmental variability, which are not present in smooth benchmark functions. We introduced a Gaussian noise into our statistical (GP) model, simulating real-life uncertainties and thereby enhancing the robustness of our proposed approach.

\item \textbf{Time and Cost Constraints}: Real-world experiments are often constrained by time and cost, whereas the cost of running the benchmark functions is very low, to the degree that it can be neglected.  To reflect the reality that the cost of running experiments is not negligible, we imposed an upper limit on the number of allowable experiments and used that threshold for evaluating the performance of our model and the competing approaches.

\item \textbf{Delays in Response}: Unlike the near instantaneous responses in the case of benchmark functions, there can be delays, sometimes substantial ones, in real-life experiments for characterizing and measuring material properties. We acknowledge that our current study did not take the delays in response into full consideration.  As such, our current study is more suitable to a process of slow changes, for which a delay in response does not significantly affect the decision outcomes. 

\item \textbf{Constraints on Experimental Combinations}: Real-life settings often involve intricate interdependencies between process variables and system behaviors, which may sometimes prevent the immediate testing of certain input combinations based on the previous experimental settings. While such complexities are challenging to replicate in simulations, our approach can indeed accommodate them, should they arise in an experiment, by treating them as new design constraints. Our use of the candidate selection sets can easily incorporate various kinds of constraints, however complex they may be, and thus provides the necessary flexibility when required.

\item \textbf{Non-Stationary Behavior}: Real-life experiments may exhibit unpredictable behaviors, such as drift in or wear-and-tear of particular equipment. These non-stationary behaviors can affect the performance of the proposed approach. While our statistical model does not consider these irregular behaviors in its current version, it can still handle these scenarios by incorporating non-i.i.d. noise and non-stationary kernel functions when necessary.  The very nature of a sequential experiment as described in this study is also an effective way to safeguard exploration under non-stationary behavior, as long as the sampling dynamics, measured by the gap between two experiments in sequence, is sufficiently faster than the drifting dynamics of the underlying process. 

\end{enumerate}

\section{Conclusions}\label{sec:conclusion}

In this work, we propose a surprise-reacting policy for guiding sequential experiments, which could be valuable for futuristic autonomous experimentation platforms. This policy dynamically switches between exploration and exploitation based on the degree of surprise, achieving a natural balance between the two and demonstrating adaptability. We show that the proposed surprise-reacting approach is effective for the rapid characterization of an unknown yet complex underlying response surface. We test the surprise-based policy using two existing surprise measures: Shannon surprise and Bayesian surprise. Our results indicate that Shannon surprise produces a faster response, aligning well with its design.

The comparison with the various acquisition functions reinforces our claim regarding the merit of conducting surprise-reacting exploitation in sequential learning. Under a resource-constrained environment, it is not effective to approximate the underlying function using more exploration-oriented approaches; nor is it effective, either, using over-exploitation approaches like EI. 

Our findings suggest that the surprise-reacting policy is more adaptive than traditional acquisition functions. A significant challenge with the use of existing acquisition functions is to determine the optimal balance between exploration and exploitation ahead of time. In contrast, the surprise-reacting policy assesses surprise observations as the sequential experiment progresses, providing an adaptive mechanism that adjusts in real time based on observed data. This adaptivity is a key advantage, enabling more efficient and effective experimentation in dynamic and uncertain environments. Our experiments also demonstrate that the rate of learning for the surprise-based approaches is faster as compared to the alternative methods, indicating a more efficient convergence to the true response surface. Overall, the surprise-reacting policy performs better in quickly approximating the response within limited evaluations, making it ideal for autonomous experimentation.

In future work, we plan to pursue several directions. First, we aim to incorporate surprise measures into an autonomous platform to conduct experiments based on its suggestions and validate the results with real experiments. We will also explore non-stationary kernels for Gaussian Processes and non-IID noise to enhance model flexibility and robustness. Additionally, we intend to develop new surprise measures beyond those used in this work and potentially extend this framework into a bandit setting to further improve adaptive experimentation strategies.

\bibliographystyle{IEEEtran}
\bibliography{TASE-SurpriseSep19}

\begin{thebibliography}{10}
\providecommand{\url}[1]{#1}
\csname url@samestyle\endcsname
\providecommand{\newblock}{\relax}
\providecommand{\bibinfo}[2]{#2}
\providecommand{\BIBentrySTDinterwordspacing}{\spaceskip=0pt\relax}
\providecommand{\BIBentryALTinterwordstretchfactor}{4}
\providecommand{\BIBentryALTinterwordspacing}{\spaceskip=\fontdimen2\font plus
\BIBentryALTinterwordstretchfactor\fontdimen3\font minus \fontdimen4\font\relax}
\providecommand{\BIBforeignlanguage}[2]{{%
\expandafter\ifx\csname l@#1\endcsname\relax
\typeout{** WARNING: IEEEtran.bst: No hyphenation pattern has been}%
\typeout{** loaded for the language `#1'. Using the pattern for}%
\typeout{** the default language instead.}%
\else
\language=\csname l@#1\endcsname
\fi
#2}}
\providecommand{\BIBdecl}{\relax}
\BIBdecl

\bibitem{nikolaev:2016}
P.~Nikolaev, D.~Hooper, F.~Webber, R.~Rao, K.~Decker, M.~Krein, J.~Poleski, R.~Barto, and B.~Maruyama, ``Autonomy in materials research: A case study in carbon nanotube growth,'' \emph{npj Computational Materials}, vol.~2, no.~1, pp. 1--6, 2016.

\bibitem{talapatra:2018}
A.~Talapatra, S.~Boluki, T.~Duong, X.~Qian, E.~Dougherty, and R.~Arr{\'o}yave, ``Autonomous efficient experiment design for materials discovery with {B}ayesian model averaging,'' \emph{Physical Review Materials}, vol.~2, no.~11, p. 113803, 2018.

\bibitem{burger:2020}
B.~Burger, P.~M. Maffettone, V.~V. Gusev, C.~M. Aitchison, Y.~Bai, X.~Wang, X.~Li, B.~M. Alston, B.~Li, R.~Clowes, N.~Rankin, B.~Harris, R.~S. Sprick, and A.~I. Cooper, ``A mobile robotic chemist,'' \emph{Nature}, vol. 583, no. 7815, pp. 237--241, 2020.

\bibitem{cao:2020}
H.~Cao, X.~Yang, and R.~Deng, ``Ontology-based holonic event-driven architecture for autonomous networked manufacturing systems,'' \emph{IEEE Transactions on Automation Science and Engineering}, vol.~18, no.~1, pp. 205--215, 2020.

\bibitem{deneault:2021}
J.~R. Deneault, J.~Chang, J.~Myung, D.~Hooper, A.~Armstrong, M.~Pitt, and B.~Maruyama, ``Toward autonomous additive manufacturing: Bayesian optimization on a 3d printer,'' \emph{MRS Bulletin}, vol.~46, no.~7, pp. 566--575, 2021.

\bibitem{lovell:2011}
C.~Lovell, G.~Jones, S.~R. Gunn, and K.-P. Zauner, ``Autonomous experimentation: Active learning for enzyme response characterisation,'' in \emph{Proceedings of Machine Learning Research, Workshop on Active Learning and Experimental Design}, vol.~16.\hskip 1em plus 0.5em minus 0.4em\relax JMLR, 2011, pp. 141--155.

\bibitem{frazier:2018}
P.~I. Frazier, ``A tutorial on {B}ayesian optimization,'' \emph{arXiv preprint arXiv:1807.02811}, 2018.

\bibitem{bull:2011}
A.~D. Bull, ``Convergence rates of efficient global optimization algorithms.'' \emph{Journal of Machine Learning Research}, vol.~12, no.~10, pp. 2879--2904, 2011.

\bibitem{chen:2019}
Z.~Chen, S.~Mak, and C.~J. Wu, ``A hierarchical expected improvement method for {B}ayesian optimization,'' \emph{Journal of the American Statistical Association}, vol. 119, no. 546, pp. 1619--1632, 2024.

\bibitem{plataniotis:2022}
N.~Plataniotis, ``Are you surprised? {T}he role of contextual surprise in designing autonomous systems,'' in \emph{IEEE 21st International Conference on Cognitive Informatics \& Cognitive Computing}, 2022, pp. 3--3.

\bibitem{baldi:2002}
P.~Baldi, ``A computational theory of surprise,'' in \emph{Information, coding and mathematics: Proceedings of workshop honoring {P}rof. {B}ob {M}celiece on his 60th birthday}.\hskip 1em plus 0.5em minus 0.4em\relax Springer, 2002, pp. 1--25.

\bibitem{itti:2006}
L.~Itti and P.~Baldi, ``Bayesian surprise attracts human attention,'' \emph{Advances in Neural Information Processing Systems}, vol.~18, 2005.

\bibitem{faraji:2018}
M.~Faraji, K.~Preuschoff, and W.~Gerstner, ``Balancing new against old information: The role of puzzlement surprise in learning,'' \emph{Neural Computation}, vol.~30, no.~1, pp. 34--83, 2018.

\bibitem{shahriari2015taking}
B.~Shahriari, K.~Swersky, Z.~Wang, R.~P. Adams, and N.~De~Freitas, ``Taking the human out of the loop: A review of bayesian optimization,'' \emph{Proceedings of the IEEE}, vol. 104, no.~1, pp. 148--175, 2015.

\bibitem{jin:2022}
S.~Jin, J.~R. Deneault, B.~Maruyama, and Y.~Ding, ``Autonomous experimentation systems and benifit of surprise-based {Bayesian} optimization,'' in \emph{Proceedings of the ISFA 2022 International Symposium on Flexible Automation}, 2022, pp. 173--179.

\bibitem{fisher:1935}
R.~A. Fisher, \emph{The Design of Experiments}.\hskip 1em plus 0.5em minus 0.4em\relax Oliver \& Boyd, Edinburgh, UK, 1935.

\bibitem{wu:2009}
C.~J. Wu and M.~S. Hamada, \emph{Experiments: Planning, Analysis, and Optimization}.\hskip 1em plus 0.5em minus 0.4em\relax John Wiley \& Sons, 2009.

\bibitem{wald:1947}
A.~Wald, \emph{Sequential Analysis.}\hskip 1em plus 0.5em minus 0.4em\relax John Wiley \& Sons, 1947.

\bibitem{box:1951}
G.~Box and K.~Wilson, ``On the experimental attainment of optimum conditions,'' \emph{Journal of the Royal Statistical Society. Series B}, vol.~13, pp. 1--45, 1951.

\bibitem{feldbaum:1960}
A.~A. Feldbaum, ``Dual control theory, {Part I},'' \emph{Automation and Remote Control}, vol.~21, no.~9, pp. 874--880, 1961.

\bibitem{cressie:1991}
N.~Cressie, \emph{Statistics for Spatial Data}.\hskip 1em plus 0.5em minus 0.4em\relax John Wiley \& Sons, 1991.

\bibitem{santner:2003}
T.~J. Santner, B.~J. Williams, W.~I. Notz, and B.~J. Williams, \emph{The Design and Analysis of Computer Experiments}.\hskip 1em plus 0.5em minus 0.4em\relax Springer, 2003.

\bibitem{kleijnen:2008}
J.~P.~C. Kleijnen, \emph{Design and Analysis of Simulation Experiments}.\hskip 1em plus 0.5em minus 0.4em\relax Springer, 2008.

\bibitem{noack:2020}
M.~M. Noack, G.~S. Doerk, R.~Li, J.~K. Streit, R.~A. Vaia, K.~G. Yager, and M.~Fukuto, ``Autonomous materials discovery driven by {Gaussian} process regression with inhomogeneous measurement noise and anisotropic kernels,'' \emph{Scientific Reports}, vol.~10, pp. 1--16, 2020.

\bibitem{chen:2021}
J.~Chen, L.~Kang, and G.~Lin, ``Gaussian process assisted active learning of physical laws,'' \emph{Technometrics}, vol.~63, no.~3, pp. 329--342, 2021.

\bibitem{chen:2022}
J.~Chen, S.~Mak, V.~R. Joseph, and C.~Zhang, ``Adaptive design for {Gaussian} process regression under censoring,'' \emph{The Annals of Applied Statistics}, vol.~16, no.~2, pp. 744--764, 2022.

\bibitem{rasmussen:2006}
C.~E. Rasmussen and C.~K. Williams, \emph{Gaussian Processes for Machine Learning}.\hskip 1em plus 0.5em minus 0.4em\relax MIT Press, 2006.

\bibitem{jones:1998}
D.~R. Jones, M.~Schonlau, and W.~J. Welch, ``Efficient global optimization of expensive black-box functions,'' \emph{Journal of Global Optimization}, vol.~13, no.~4, pp. 455--492, 1998.

\bibitem{kushner:1964}
H.~J. Kushner, ``A new method of locating the maximum point of an arbitrary multipeak curve in the presence of noise,'' \emph{Journal of Basic Engineering}, vol.~86, no.~1, pp. 97--106, 1964.

\bibitem{cox:1992}
D.~D. Cox and S.~John, ``A statistical method for global optimization,'' in \emph{Proceedings of the 1992 IEEE International Conference on Systems, Man, and Cybernetics}.\hskip 1em plus 0.5em minus 0.4em\relax IEEE, 1992, pp. 1241--1246.

\bibitem{srinivas:2010}
N.~Srinivas, A.~Krause, S.~Kakade, and M.~Seeger, ``Gaussian process optimization in the bandit setting: No regret and experimental design,'' in \emph{Proceedings of the 27th International Conference on Machine Learning (ICML-10)}, 2010, pp. 1015--1022.

\bibitem{mockus:1978}
J.~Mockus, V.~Tiesis, and A.~Zilinskas, ``Application of {Bayesian} approach to numerical methods of global and stochastic optimization,'' \emph{Journal of Global Optimization}, vol.~4, pp. 117--129, 1978.

\bibitem{gongora:2020}
A.~E. Gongora, B.~Xu, W.~Perry, C.~Okoye, P.~Riley, K.~G. Reyes, E.~F. Morgan, and K.~A. Brown, ``A {B}ayesian experimental autonomous researcher for mechanical design,'' \emph{Science Advances}, vol.~6, p. eaaz1708, 2020.

\bibitem{albahar:2021}
A.~Al~Bahar, I.~Kim, and X.~Yue, ``A robust asymmetric kernel function for {Bayesian} optimization, with application to image defect detection in manufacturing systems,'' \emph{IEEE Transactions on Automation Science and Engineering}, vol.~19, no.~4, pp. 3222--3233, 2021.

\bibitem{kang:2023}
B.~Kang, C.~Park, H.~Kim, and S.~Hong, ``Bayesian optimization for the vehicle dwelling policy in a semiconductor wafer fab,'' \emph{IEEE Transactions on Automation Science and Engineering}, online published, 2023.

\bibitem{lancaster:2018}
J.~Lancaster, R.~Lorenz, R.~Leech, and J.~H. Cole, ``Bayesian optimization for neuroimaging pre-processing in brain age classification and prediction,'' \emph{Frontiers in Aging Neuroscience}, vol.~10, p.~28, 2018.

\bibitem{zhang:2020}
Y.~Zhang, D.~W. Apley, and W.~Chen, ``Bayesian optimization for materials design with mixed quantitative and qualitative variables,'' \emph{Scientific Reports}, vol.~10, no.~1, pp. 1--13, 2020.

\bibitem{wang2023recent}
X.~Wang, Y.~Jin, S.~Schmitt, and M.~Olhofer, ``Recent advances in {Bayesian} optimization,'' \emph{ACM Computing Surveys}, vol.~55, no. 13s, pp. 1--36, 2023.

\bibitem{ghorbani2024active}
M.~Ghorbani, M.~Boley, P.~Nakashima, and N.~Birbilis, ``An active machine learning approach for optimal design of magnesium alloys using {Bayesian} optimisation,'' \emph{Scientific Reports}, vol.~14, no.~1, p. 8299, 2024.

\bibitem{di2024active}
F.~Di~Fiore, M.~Nardelli, and L.~Mainini, ``Active learning and {Bayesian} optimization: {A} unified perspective to learn with a goal,'' \emph{Archives of Computational Methods in Engineering}, pp. 1--29, 2024.

\bibitem{han2023adaptive}
G.~Han, J.~Jeong, and J.-H. Kim, ``Adaptive {Bayesian} optimization for fast exploration under safety constraints,'' \emph{IEEE Access}, vol.~11, pp. 42\,949--42\,969, 2023.

\bibitem{islam2024dynamic}
U.~J. Islam, K.~Paynabar, G.~Runger, and A.~S. Iquebal, ``Dynamic exploration-exploitation trade-off in active learning regression with {Bayesian} hierarchical modeling,'' \emph{IISE Transactions}, online published, 2024.

\bibitem{bukkapatnam:2022}
S.~T. Bukkapatnam, ``Autonomous materials discovery and manufacturing ({AMDM}): A review and perspectives,'' \emph{IISE Transactions}, vol.~55, no.~1, pp. 75--93, 2023.

\bibitem{sobol1967}
I.~M. Sobol', ``On the distribution of points in a cube and the approximate evaluation of integrals,'' \emph{Zhurnal Vychislitel'noi Matematiki i Matematicheskoi Fiziki}, vol.~7, no.~4, pp. 784--802, 1967.

\bibitem{johnson:1990}
M.~E. Johnson, L.~M. Moore, and D.~Ylvisaker, ``Minimax and maximin distance designs,'' \emph{Journal of Statistical Planning \& Inference}, vol.~26, pp. 131--148, 1990.

\bibitem{morris:1995}
M.~D. Morris and T.~J. Mitchell, ``Exploratory designs for computational experiments,'' \emph{Journal of Statistical Planning \& Inference}, vol.~43, pp. 381--402, 1995.

\bibitem{joseph:2015}
V.~R. Joseph, E.~Gul, and S.~Ba, ``Maximum projection designs for computer experiments,'' \emph{Biometrika}, vol. 102, no.~2, pp. 371--380, 2015.

\bibitem{omohundro1989}
S.~M. Omohundro, ``Five balltree construction algorithms,'' \emph{Technical Report of International Computer Science Institute, TR-89-063}, 1989.

\bibitem{makela:2017}
M.~M{\"a}kel{\"a}, ``Experimental design and response surface methodology in energy applications: A tutorial review,'' \emph{Energy Conversion and Management}, vol. 151, pp. 630--640, 2017.

\bibitem{habib:2009}
S.~S. Habib, ``Study of the parameters in electrical discharge machining through response surface methodology approach,'' \emph{Applied Mathematical Modelling}, vol.~33, no.~12, pp. 4397--4407, 2009.

\bibitem{liang:2021}
Q.~Liang, A.~E. Gongora, Z.~Ren, A.~Tiihonen, Z.~Liu, S.~Sun, J.~R. Deneault, D.~Bash, F.~Mekki-Berrada, S.~A. Khan \emph{et~al.}, ``Benchmarking the performance of bayesian optimization across multiple experimental materials science domains,'' \emph{npj Computational Materials}, vol.~7, no.~1, p. 188, 2021.

\bibitem{de2015}
S.~G. de-los Cobos-Silva, M.~{\'A}. Guti{\'e}rrez-Andrade, R.~A. Mora-Guti{\'e}rrez, P.~Lara-Vel{\'a}zquez, E.~A. Rinc{\'o}n-Garc{\'\i}a, and A.~Ponsich, ``An efficient algorithm for unconstrained optimization,'' \emph{Mathematical Problems in Engineering}, vol. 2015, no.~1, p. 178545, 2015.

\bibitem{lee2020}
E.~Lee, D.~Eriksson, D.~Bindel, B.~Cheng, and M.~Mccourt, ``Efficient rollout strategies for bayesian optimization,'' in \emph{Proceedings of the 36th Conference on Uncertainty in Artificial Intelligence}, vol. 124.\hskip 1em plus 0.5em minus 0.4em\relax PMLR, 2020, pp. 260--269.

\bibitem{botcha:2021}
B.~Botcha, A.~S. Iquebal, and S.~T. Bukkapatnam, ``Efficient manufacturing processes and performance qualification via active learning: Application to a cylindrical plunge grinding platform,'' \emph{Procedia Manufacturing}, vol.~53, pp. 716--725, 2021.

\bibitem{sanjeevi:2021}
R.~Sanjeevi, G.~A. Kumar, and B.~R. Krishnan, ``Optimization of machining parameters in plane surface grinding process by response surface methodology,'' \emph{Proceedings of Materials Today}, vol.~37, pp. 85--87, 2021.

\end{thebibliography}

%
\vspace{-1.5 cm}

\begin{IEEEbiography}[{\includegraphics[width=1in,height=1.25in,clip,keepaspectratio]{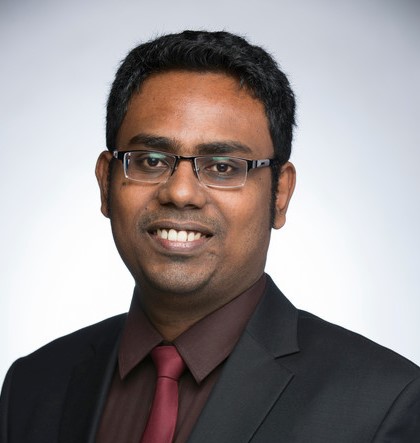}}]{Imtiaz Ahmed}
received B.Sc. and M.Sc. in Industrial \& Production Engineering from Bangladesh University of Engineering \& Technology, Dhaka, Bangladesh in 2012 and 2014 respectively. He received his PhD degree in Industrial Engineering from Texas A\&M University in 2020. He is currently an Assistant Professor of Industrial \& Management Systems Engineering Department at the West Virginia University. His research interests are in data and decision science.
\end{IEEEbiography}

\vspace{-2 cm}

\begin{IEEEbiography}[{\includegraphics[width=1in,height=1.25in,clip,keepaspectratio]{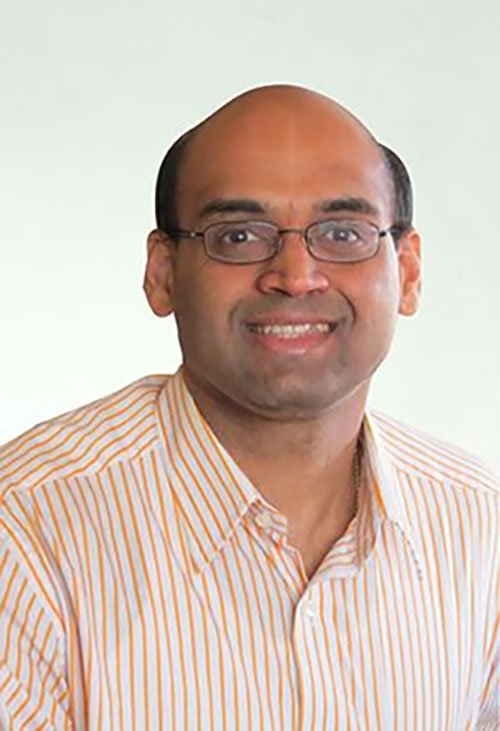}}]{Satish T.S Bukkapatnam} is the Rockwell International Professor of Industrial \& Systems Engineering at Texas A\&M University. He received his PhD degree in Industrial and Manufacturing Engineering from Pennsylvania State University (1997). His research interests are broadly in smart manufacturing systems, and ultraprecision manufacturing. Dr. Bukkapatnam is a Fellow of IISE and SME, Associate Member of CIRP, and was a Fulbright-Tocqueville Distinguished Chair.
\end{IEEEbiography}

\vspace{-2 cm}
\begin{IEEEbiography}[{\includegraphics[width=1in,height= 1.25in,clip,keepaspectratio]{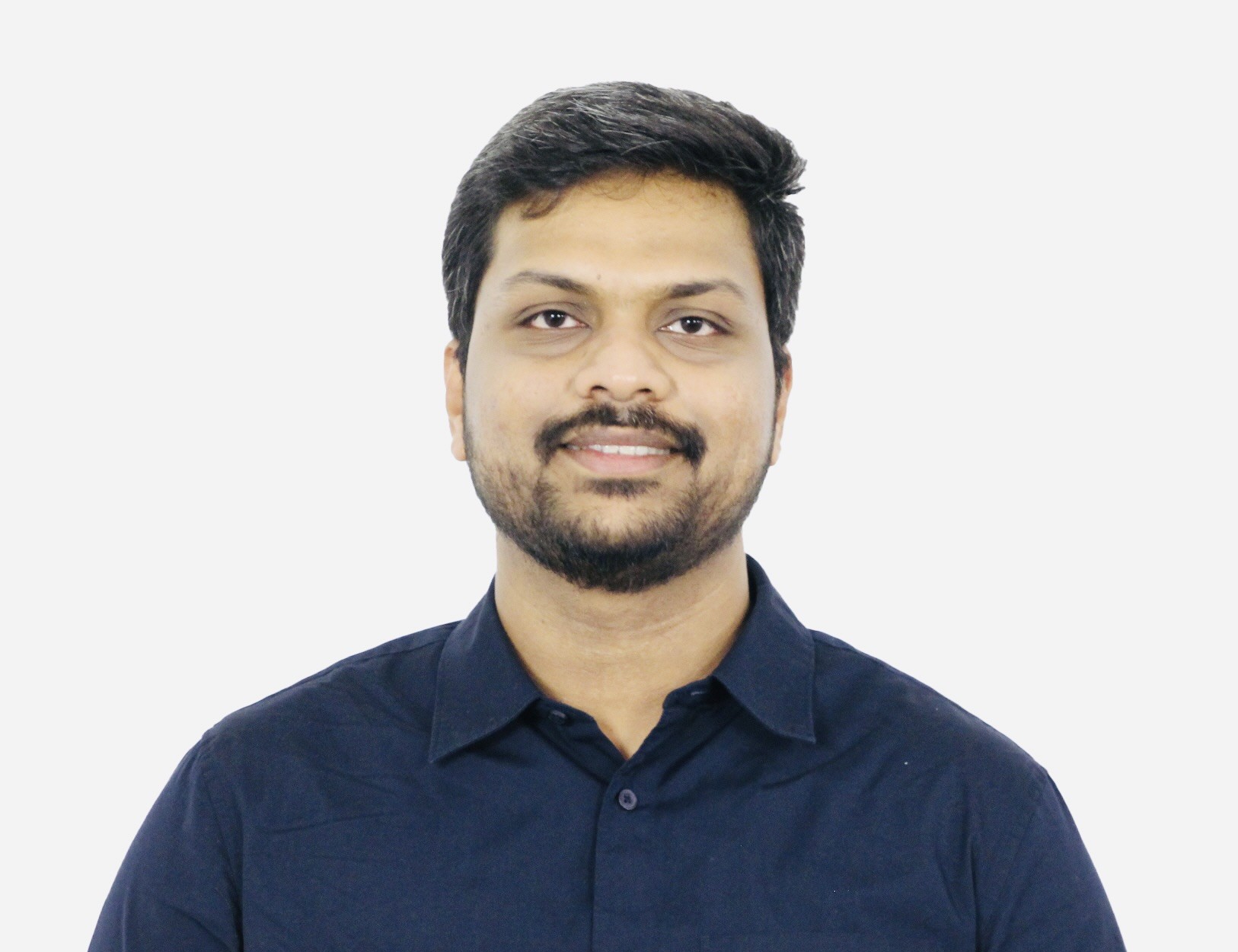}}]{Bhaskar Botcha} received his B.Tech in Mechanical Engineering and M.Tech from Indian Institute of Technology, Madras, in 2017. He is currently working towards his Ph.D. degree in Industrial \& Systems Engineering at Texas A\&M University. He previously was an intern at the Lawrence Livermore National Laboratory. His research interests include smart manufacturing, Industry 4.0 and using machine learning and signal processing techniques for advanced materials discovery.
\end{IEEEbiography}

\vspace{-2 cm}
\begin{IEEEbiography}[{\includegraphics[width=1in,height=1.25in,clip,keepaspectratio]{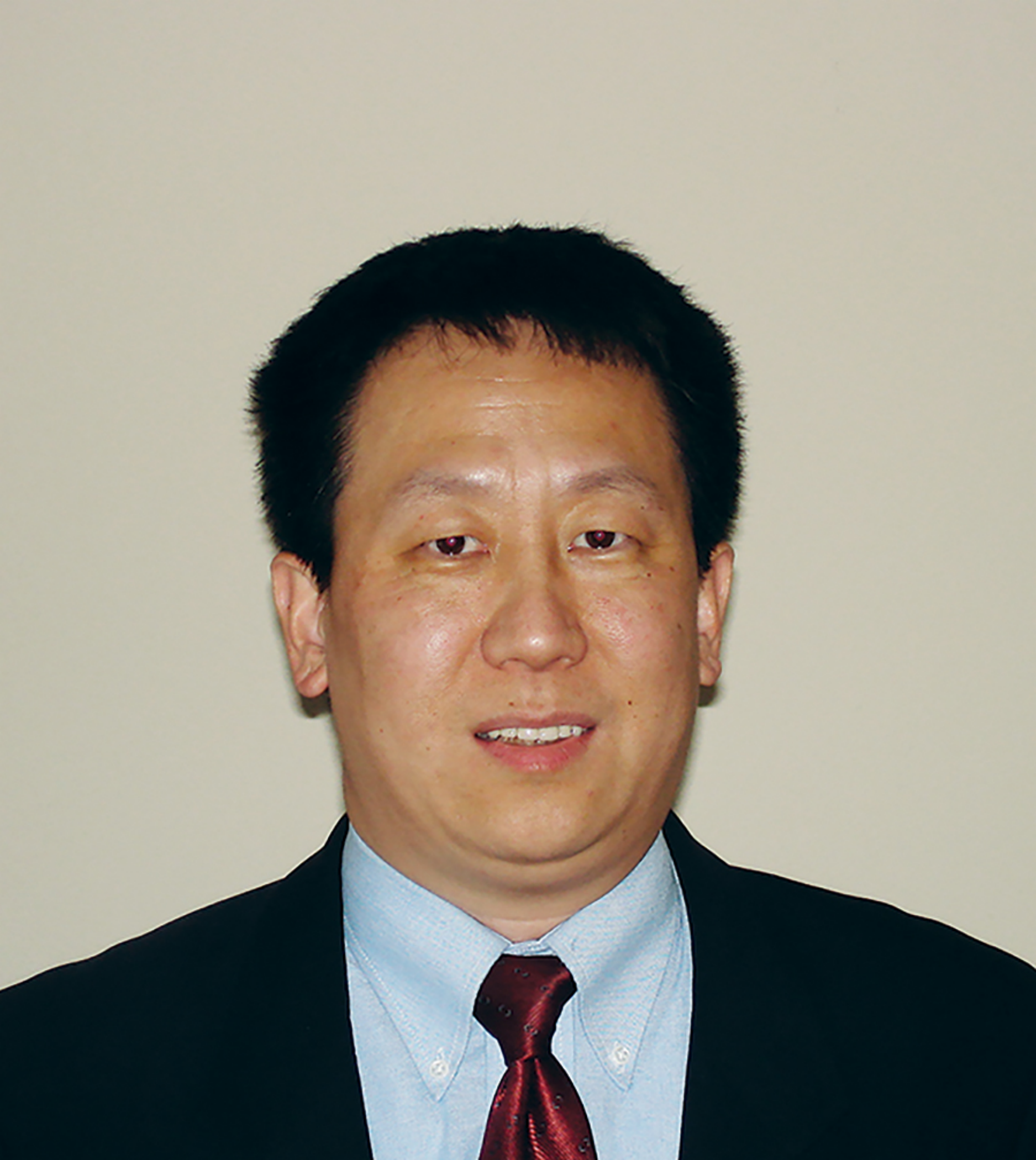}}]{Yu Ding}
(M'01, SM'11) received B.S. from University of Science \& Technology of China (1993); M.S. from Tsinghua University, China (1996); M.S. from Penn State University (1998); received Ph.D. in Mechanical Engineering from University of Michigan (2001). He is currently Anderson-Interface Chair and Professor at the H. Milton Stewart School of Industrial \& Systems Engineering at Georgia Institute of Technology. Prior to joining Georgia Tech, he was the Mike and Sugar Barnes Professor at Texas A\&M University. His research interests are in data and quality science. Dr. Ding is the Editor-in-Chief of \emph{IISE Transactions} for the term of 2021--2024. Dr. Ding is a fellow of IIE, a fellow of ASME, a senior member of IEEE, and a member of INFORMS.
\end{IEEEbiography}


\end{document}